\pdfoutput=1
\documentclass[11pt]{article}

\usepackage{acl}

\usepackage{times}
\usepackage{latexsym}

\usepackage[T1]{fontenc}
\usepackage[utf8]{inputenc}

\usepackage{microtype}

\usepackage{inconsolata}

\usepackage{graphicx}
\usepackage{booktabs}
\usepackage{adjustbox}
\usepackage{colortbl}
\usepackage{multirow}
\usepackage{pifont}

\usepackage{amssymb}
\usepackage{amsmath}
\usepackage{bm}

\definecolor{yellow-green}{rgb}{0.6, 0.8, 0.2}
\definecolor{bluegray}{rgb}{0.4, 0.6, 0.8}
\definecolor{orange-red}{rgb}{1.0, 0.27, 0.0}

\usepackage[linesnumbered,ruled,vlined]{algorithm2e}

\SetCommentSty{mycommfont}

\usepackage{afterpage}
\usepackage{here}

\newcommand{\Add}[1]{\textcolor{black}{#1}}

\title{DiLM: Distilling Dataset into Language Model \\for Text-level Dataset Distillation}

\newcommand{\mail}[2]{\href{mailto:#1}{\color{black}{#2}}}
\author{
    Aru Maekawa\quad
    Satoshi Kosugi\quad
    Kotaro Funakoshi\quad
    Manabu Okumura \\
    Tokyo Institute of Technology\\
    \texttt{%
        \{%
          \mail{maekawa@lr.pi.titech.ac.jp}{maekawa},
          \mail{kosugi@lr.pi.titech.ac.jp}{kosugi},
          \mail{funakoshi@lr.pi.titech.ac.jp}{funakoshi},
          \mail{oku@lr.pi.titech.ac.jp}{oku}%
        \}%
        @lr.pi.titech.ac.jp
    }
}

\begin{document}
\maketitle

\begin{abstract}
Dataset distillation aims to compress a training dataset by creating a small number of informative synthetic samples such that neural networks trained on them perform as well as those trained on the original training dataset.
% In this paper, we explore the question of whether we can train a language model to generate more effective training data than the real samples of the original training dataset for the text dataset distillation.
Current text dataset distillation methods create each synthetic sample as a sequence of word embeddings instead of a text to apply gradient-based optimization; however, such embedding-level distilled datasets cannot be used for training other models whose word embedding weights are different from the model used for distillation.
To address this issue, we propose a novel text dataset distillation approach, called \emph{Distilling dataset into Language Model (DiLM)}, which trains a language model to generate informative synthetic training samples as text data, instead of directly optimizing synthetic samples.
We evaluated DiLM on various text classification datasets and showed that distilled synthetic datasets from DiLM outperform those from current coreset selection methods. DiLM achieved remarkable generalization performance in training different types of models and in-context learning of large language models.
% the representative real samples of the original datasets.
% Moreover, our text-level distilled datasets of DiLM show the remarkable generalization performance for training other pre-trained models that have different word embeddings, for which embedding-level distilled datasets are completely useless, and in-context learning of LLMs.
Our code will be available at \url{https://github.com/arumaekawa/DiLM}.
\end{abstract}

\section{Introduction}

% 大規模モデルの学習コストの問題
The successful advancements in machine learning in a wide range of fields are due to the scaling-up of deep neural networks and large training datasets. 
In the natural language processing (NLP) field, large language models (LLMs), which are pre-trained with a huge amount of text, such as BERT- and GPT-family models~\cite{devlin-etal-2019-bert,liu-2019-roberta,radford2019language,brown2020gpt3}, have shown remarkable capabilities for various NLP tasks. 
However, training such large-scale models requires large computational resources and a long time, which makes it difficult to develop new LLMs, and even to fine-tune them.

% 提案法の概略図
\begin{figure*}[t]
    \centering
    \includegraphics[width=\linewidth]{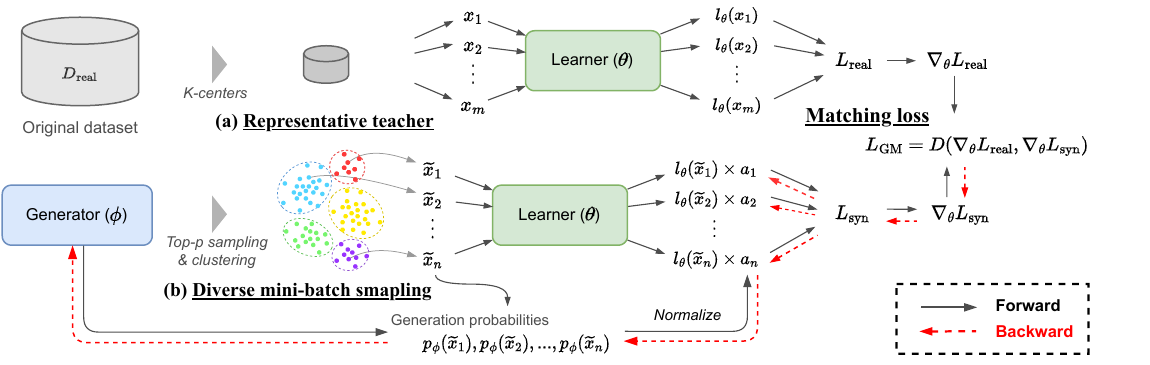}
    \caption{Overview of training with DiLM. Gradient matching loss is computed on the learner model between real samples from the original dataset and generated samples from the generator model. It is then back-propagated to the generator model via generation probabilities, which weight the learner loss for each generated sample. (a) Representative teacher for computing real sample's gradients, which improves the performance and accelerates convergence by using K-center samples, representing the original dataset, rather than randomly sampled ones. (b) Diverse mini-batch sampling, which enables the generator model to explore diverse synthetic samples in each training step.}
    \label{fig:overview}
\end{figure*}

% データセット蒸留
To address this issue, dataset distillation~\cite{wang2018dataset} has attracted much attention in the machine learning community, which aims to reduce training costs by compressing training datasets. 
In contrast to traditional coreset selection approaches~\cite{wolf2011facility,sener2018active,welling2009herding}, which heuristically select a small subset of representative training samples from the original dataset, dataset distillation creates more informative synthetic samples by distilling the knowledge from the original dataset. 
With this approach, synthetic samples are optimized with gradient descent according to objective functions for dataset distillation, including meta-learning~\cite{wang2018dataset}, gradient matching~\cite{zhao2021dataset}, training trajectory matching~\cite{Cazenavette_2022_CVPR}, and feature distribution matching~\cite{wang2022cafe,zhao2023distribution}.
The recent remarkable performance of dataset distillation, especially in the computer vision (CV) field, has also led to studies of its various applications, including neural architecture search~\citep{pmlr-v119-such20a,medvedev2022learning}, federated learning~\citep{zhang2022dense,xiong2023feddm}, continual learning~\citep{wiewel-2021-condensed-conposite-memory,sangermano-2022-sample-condensation}, and privacy preservation~\citep{pmlr-v162-dong22c,chen-2022-private-set}. 

% (今回解決したい課題) テキストは離散なので単語埋め込みとして最適化するけどそれだと問題がある
While most previous studies applied dataset distillation only to image classification datasets, some studies focused on text dataset distillation~\cite{sucholutsky-2021-soft-label,li-2021-data-distillation,maekawa-etal-2023-dataset,sahni2023multilingual}. 
% In contrast to the image, which can be considered as a pixel-wise continuous data to apply the gradient-based optimization, the discrete nature of text makes the dataset distillation challenging problem~\cite{Geng2023survey,yu2023review}. 
In contrast to the image, which can be applied gradient-based optimization by considering it as a pixel-wise continuous data, the discrete nature of text makes dataset distillation challenging~\cite{Geng2023survey,yu2023review}. 
To address this issue, all existing text dataset distillation methods used the widely used neural NLP technique called embedding, i.e., optimizing a synthetic dataset as continuous input word embeddings instead of discrete text. 
However, such embedding-level distilled synthetic datasets cannot be used for training other models that have different word embedding weights, which is a crucial issue in terms of practical applications. 
Furthermore, distilled word embedding sequences are also completely unreadable to humans, which makes it difficult to interpret and analyze the original training dataset by observing distilled synthetic samples. 

% (提案手法) 離散的なテキストを最適化する代わりにテキスト生成モデルを最適化する
% In this paper, we make two contributions. 
% Firstly, we tackle the problem of text dataset distillation to obtain distilled synthetic datasets in text-level as the first work for model-agnostic applicability and interpretability.
% Secondly, to achive this goal, we propose the first text-level dataset distillation approach called ``\textbf{D}istill dataset \textbf{i}nto \textbf{L}anguage \textbf{M}odel (\textbf{DiLM})''.
Motivated by these shortcomings, this paper explores the text dataset distillation to obtain distilled synthetic datasets at the text-level as the first study. 
We propose the first text-level dataset distillation approach called ``\textbf{D}istilling dataset \textbf{i}nto \textbf{L}anguage \textbf{M}odel (\textbf{DiLM})''. 
To overcome the optimization difficulty of discrete text, DiLM uses a language model as a surrogate continuous optimization target instead of directly optimizing a synthetic sample's text.
Specifically, DiLM trains a language model to minimize the gradient matching loss~\cite{zhao2021dataset} of generated synthetic samples as a dataset distillation objective.
To enable back-propagating the gradient matching loss to the language model, we design a differentiable backward pass via loss weighting with generation probabilities to bypass the non-differentiable generated text (Figure~\ref{fig:overview}).

% (実験結果)
In our experiments, we applied DiLM to distill three text classification datasets from the GLUE benchmark~\cite{wang-etal-2018-glue}, SST-2, QQP, and MNLI-m.
The results indicate that the synthetic datasets distilled with DiLM outperformed representative real samples selected from the original datasets with current %state-of-the-art (SOTA) 
coreset selection methods. 
Our distilled datasets also achieved remarkable generalization performance not only for training different types of pre-trained models but also for in-context learning of LLMs as few-shot prompts.

% 貢献点を列挙しておく
Our main contributions are as follows:
\begin{itemize}
    \setlength\itemsep{0em}
    % テキストとして蒸留データセットを獲得する
    \item To the best of our knowledge, this is the first study to distill a text dataset into a text-level synthetic dataset that are applicable for training models independent of word embedding weights.
    % テキストの離散性への対処
    \item We present DiLM, which addresses the discreteness of text by using a language model as a surrogate optimization target and back-propagating the distillation loss to the model, bypassing non-differentiable generated text.
    % モデルによらずコアセットより良い性能を達成
    \item Our experimental results indicate that DiLM outperformed the current coreset selection methods not only for training the same model used for distillation, but also for training different models independent of the word embedding weights, architectures, and training processes.
\end{itemize}

\section{Related Work}

\subsection{Dataset Distillation}
Dataset distillation was first proposed by \citet{wang2018dataset}, motivated by theoretical interests as well as practical applications for reducing network training costs. 
Inspired by meta-learning based hyperparameter optimization~\cite{maclaurin2015gradient}, \citet{wang2018dataset} optimized a small synthetic dataset by gradient descent such that models trained on it have a lower training loss for the original dataset.
Recently, several surrogate objectives have been proposed to improve the performance and efficiency of dataset distillation.
DC~\cite{zhao2021dataset} and DSA~\cite{pmlr-v139-zhao21a} focused on gradient matching between real and synthetic samples.
DM~\cite{zhao2023distribution} and CAFE~\cite{wang2022cafe} proposed feature distribution matching, which requires less GPU memory for optimizing synthetic datasets.
MTT~\cite{Cazenavette_2022_CVPR} and TESLA~\cite{pmlr-v202-cui23e} optimized synthetic samples to approximate trajectories of model parameters trained with real data. 
SLDD~\cite{sucholutsky-2021-soft-label} and LDD~\cite{bohdal-2020-flexible-dataset} introduced learnable soft-labels, which are optimized together with input images to make each synthetic sample more informative.

While the most current research on dataset distillation involves only image classification datasets, some studies also focused on text classification datasets. 
\citet{sucholutsky-2021-soft-label} and \citet{li-2021-data-distillation} applied the original meta-learning based method by \citet{wang2018dataset} to text datasets. 
To overcome the discrete nature of text, which makes applying gradient-based methods difficult, they optimized synthetic samples in the pre-trained GloVe word embedding space~\cite{pennington-etal-2014-glove} instead of actual words of text as the optimization target. 
\citet{maekawa-etal-2023-dataset} extended the text dataset distillation to the pre-trained BERT model and improved its performance by introducing learnable attention labels, which directly guide the self-attention probabilities of the models.
\citet{sahni2023multilingual} explored dataset distillation in multilingual text classification datasets in the context of fairness, interpretability, and cross-architecture generalization. 
Although these methods perform well for text classification datasets, distilled synthetic datasets obtained with them cannot be used for training other models that have different word embedding weights.
Although \citet{sucholutsky-2021-soft-label} and \citet{sahni2023multilingual} transformed their distilled synthetic samples to text by finding a word that has the nearest neighbor embedding, the converted text consists of unrelated words and does not make sense, which makes it difficult to interpret and analyze them. 
Moreover, %no previous work investigated the performance of distilled datasets after converted to text.
the performance of distilled datasets after being converted to text has also not been investigated.

\subsection{Generative Models}
\label{sec:related_work/generative_models}
% Recent works also showed that Generative Adversarial Network (GAN) and GAN-Inversion are effective for the image dataset distillation.
Recent studies on dataset distillation in the CV field used generative adversarial networks (GANs)~\cite{ian2014gan}, i.e., training the model parameters and/or their latent input noises instead of synthetic images.
These methods generalize distilled synthetic images to different model architectures by restricting them to the generative distribution learned from the original dataset.
DiM~\cite{wang2023dim} fine-tuned a GAN to generate informative synthetic images from randomly sampled latent noises, where distilled datasets of different sizes can be produced without retraining the model.
GTNs~\cite{pmlr-v119-such20a} trained a GAN to generate informative images, instead of realistic images, to accelerate neural architecture search. 
GTNs also learned a latent noise for each synthetic image as a curriculum of training learner networks.
IT-GAN~\cite{zhao2022itgan} and GLaD~\cite{cazenavette2023generative} used a pre-trained GAN as a generative prior of synthetic samples and only optimized the latent noises.

Inspired by these studies, we also introduce a generative model with a different motivation for text dataset distillation: to avoid the difficulties of directly optimizing discrete text, we instead optimize the continuous parameters of a generative model to generate distilled synthetic samples.
However, since all previous studies that used generative models for image dataset distillation trained them %a generative model
and/or their input latent noises by back-propagating the distillation loss to them via generated images, none of them can be applied to text data, which are non-differentiable due to their discrete nature. 

\section{Methodology}

In this section, we introduce DiLM, which distills text datasets into text data, not word embeddings, for the model-agnostic applicability and interpretability of the distilled synthetic datasets. 
The main idea of DiLM is to avoid the optimization difficulties of discrete text by instead training continuous parameters of a language model as a surrogate optimization target of dataset distillation. 

\subsection{Overview}
Given a training dataset $\mathcal{D}_\text{real} = \{x_i\}_{i=1}^{|\mathcal{D}_\text{real}|}$, the goal of DiLM is to obtain a generator model, parameterized by $\phi$, that generates a distilled synthetic dataset $\mathcal{D}_\text{syn} = \{\tilde{x}_i\}_{i=1}^{|\mathcal{D}_\text{syn}|}$ ($|\mathcal{D}_\text{syn}| \ll |\mathcal{D}_\text{real}|$), such that a learner model, parameterized by $\theta$, trained on $\mathcal{D}_\text{syn}$ performs well.
To achieve this goal, the overall procedure of DiLM is composed of the following three steps:
\begin{enumerate}
\setlength\itemsep{0em}
%
% \noindent\textbf{1.}
\item We first simply train the generator model to generate synthetic training samples that belong to the same distribution as in the original dataset $\mathcal{D}_\text{real}$ (Section~\ref{sec:text_generation_model}). 
%
% \noindent\textbf{2.}
\item We then fine-tune the generator model to generate ``informative'' training samples by minimizing the gradient matching loss between generated and real samples (Section~\ref{sec:text_gtn}). 
%
%\noindent\textbf{3.}
\item We obtain distilled dataset $\mathcal{D}_\text{syn}$ by generating synthetic samples with the generator model and selecting representative samples from them by using a clustering-based coreset selection method (Section~\ref{sec:sample_selection}). 
\end{enumerate}
We describe the details of each step in the following sections.

%\subsection{Training Data Generation}
\subsection{Synthetic Training Data Generation with Language Model}
\label{sec:text_generation_model}

Inspired by the remarkable text generation capability of pre-trained transformer language models~\cite{radford2019language}, we use them as the generator model to generate synthetic training samples of sufficient quality to be used for training models. 
Before training the generator model to generate more informative synthetic samples than real samples in the original dataset, we first simply train a language model to generate training samples that belong to the same distribution as in the original training dataset for the initial parameters of the generator model. 

When we target at text classification tasks, we need to control the generator model to generate samples for each specific class. 
Therefore, we introduce class-specific beginning-of-sentence tokens \verb|<bos_i>|, which are added to the head of each training sample to train the generator model to generate samples of the corresponding class following it.
For each training sample, an end-of-sentence token \verb|<eos>| is also added, and the sample is fed to the generator model as follows:
\begin{quote}
    \verb|<bos_i>| \textit{sentence of class $i$} \verb|<eos>|.
\end{quote}
To involve text classification tasks that specify the relation between two sentences, such as semantic similarity and natural language inference (NLI), we use a separate token \verb|<sep>| to split two sentences as
\begin{quote}
    \verb|<bos_i>| \textit{sentence~1} \verb|<sep>| \textit{sentence~2} \verb|<eos>|.
\end{quote}

The generator model is trained on them with the language modeling loss $l_\phi(x_i)$ as
\begin{align}
    l_\phi(x_i) = - \frac{1}{|x_i|}\sum_{w_t\in x_i} \log p_\phi(w_t|w_{<t}),
\end{align}
where $w_t$ is a token in $x_i$ and $|x_i|$ is the length of $x_i$. In this way, we pre-train the generator model parameters $\phi$ to generate synthetic training data like real data, and use them as the initial parameter for training for gradient matching, described in the following section. % and as a baseline to validate the effectiveness of training DiLM.

\subsection{Training for Gradient Matching}
\label{sec:text_gtn}

In this section, we explain how to fine-tune the pre-trained generator model, described in Section~\ref{sec:text_generation_model}, to generate synthetic training samples that are more informative than real samples in the original dataset.
Specifically, we describe gradient matching, which is an optimization objective for dataset distillation, and the model updating procedure to deal with the discreteness of text.
We also introduce two techniques to improve DiLM: representative teacher and diverse mini-batch sampling.
% We summarize the overall training procedure of DiLM in Algorithm~\ref{alg:text_gtn}.

%%%%%%%%%%%%%
% Algorithm %
%%%%%%%%%%%%%

\begin{algorithm}[t]
    % \footnotesize
    \small
    \SetInd{0.4em}{1.0em}
    \DontPrintSemicolon
    \SetKwInOut{Input}{Input}
    \SetKwInOut{Output}{Output}
    \Input{
        $\mathcal{D}_\text{real}$: original dataset;
        $\phi$: generator model;
        $\theta$: learner model;
        $S$: \#~of outer loop;
        $T$: \#~of inner loop;
        $K$: \#~of learner updating loop in each inner step;
        $M$: batch size of real data;
        $N$: batch size of synthetic data;
        $\eta$: learning rate of $\theta$;
        $\alpha$: learning rate of $\phi$.
    }
    \tcp{Outer loop}
    \For{$s=1$, \dots, $S$ } {
        \tcp{Initialize learner}
        Initialize $\theta\sim p(\theta_0)$ \\
        \tcp{Inner loop}
        \For{$t=1$, \dots, $T$} {
            \tcp{Compute gradient matching loss for each class}
            \For{$c=1$, \dots, $C$} {
                \tcp{Compute loss with real samples}
                $\{x_i^{(c)}\}_{i=1}^{M}\sim \mathcal{D}_\text{real}^{(c)}$ \\
                $\mathcal{L}_\text{real}^{(c)} \leftarrow \frac{1}{M}\sum_{i=1}^{M}l_\theta(x_i^{(c)})$ \\
                \tcp{Compute loss with synthetic samples}
                $\{\tilde{x}_i^{(c)}\}_{i=1}^{N}\sim p_\phi(\tilde{x})$ \\
                \For{$i=1$, \dots, $N$} {
                    $a_i \leftarrow p_\phi(\tilde{x}_i^{(c)}) / \sum_{j=1}^N p_\phi(\tilde{x}_j^{(c)})$\\
                }
                $\mathcal{L}_\text{syn}^{(c)} \leftarrow \sum_{i=1}^{N}a_i l_\theta(\tilde{x}_i^{(c)})$ \\
                \tcp{Gradient matching loss (Eq.~\eqref{eq:distance_function})}
                $\mathcal{L}_\text{GM}^{(c)} \leftarrow  D(
                    \nabla_\theta\mathcal{L}_\text{real}^{(c)},
                    \nabla_\theta\mathcal{L}_\text{syn}^{(c)})
                )$
            }
            \tcp{Update generator}
            $\phi \leftarrow \phi - \alpha \nabla_{\phi} \frac{1}{C}\sum_{c=1}^{C}\mathcal{L}_\text{GM}^{(c)}$\\
            \tcp{Update learner for K steps}
            \For{$k=1$, \dots, $K$} {
                $X_\text{real}\sim \mathcal{D}_\text{real}$\\
                $\theta \leftarrow \theta - \eta \nabla_{\theta} \mathcal{L}_\theta(X_\text{real})$ 
            }
        }
    }
    \Output{$\phi$: parameters of generator model.}
    \caption{Optimization for DiLM}
    \label{alg:text_gtn}
\end{algorithm}

% \paragraph{Gradient matching.}
\noindent\textbf{Gradient Matching.}
To distill the knowledge of the original dataset $\mathcal{D}_\text{real}$ into generated synthetic samples from the generator model, we optimize the gradient matching loss~\cite{zhao2021dataset} as the objective for dataset distillation.
Given a mini-batch of real samples $\{x_i\}_{i=1}^M$ and a mini-batch of synthetic samples $\{\tilde{x}_i\}_{i=1}^N$, which is generated from the generator model, the gradient matching loss $\mathcal{L}_\text{GM}$ on the learner model parameters $\theta$ is calculated as
\begin{align}
    \begin{split}
    \mathcal{L}_\text{GM} &= D\left(
        \nabla_\theta \mathcal{L}_\text{real}, \nabla_\theta \mathcal{L}_\text{syn}
    \right)\label{eq:gradient_matching_loss} \quad \text{where} \\
    \mathcal{L}_\text{real} &= \frac{1}{M}\sum_{i=1}^M l_\theta(x_i), \ 
    \mathcal{L}_\text{syn} = \frac{1}{N}\sum_{i=1}^N l_\theta(\tilde{x}_i),
    \end{split}
\end{align}
where $l_\theta(\cdot)$ is the loss function for learning tasks such as cross-entropy loss, and $D(\cdot,\cdot)$ is the cosine similarity-based distance function, expressed as
\begin{align}
    D(A, B) = 1-\frac{A\cdot B}{\|A\|\|B\|}.
    \label{eq:distance_function}
\end{align}
Following a previous study~\cite{zhao2021dataset}, we separately calculate the gradient matching loss for each class and combine them to update the generator model parameters $\phi$. 
To consider the gradient on the learner model parameters $\theta$ throughout the entire training process, the generator model is trained with the nested loop algorithm, including the outer loop, which initializes $\theta$ at the beginning, and the inner loop, which updates $\theta$ for $K$ steps with real samples (see Algorithm~\ref{alg:text_gtn}).

% \paragraph{Generator updating.}
\noindent\textbf{Generator Updating.}
As we described in Section~\ref{sec:related_work/generative_models}, the gradient matching loss $\mathcal{L}_\text{GM}$ cannot be directly back-propagated to the generator model parameters $\phi$ via generated samples $\{\tilde{x}_i\}_{i=1}^N$, like the case with image datasets, because they consist of discrete text.
\Add{Although some solutions to the discrete back-propagation issue in text generation have been explored in the NLP research field, most of standard approaches, including soft-argmax~\cite{zhang-etal-2017-textgan} and policy gradient~\cite{yu-etal-2017-seqgan}, cannot be applied to this case (see details in Appendix~\ref{sec:appendix/background}).}
To address this issue, we design an alternative backward pass, inspired by a previous study~\cite{hiraoka-etal-2020-optimizing}, which optimizes a tokenization model for the downstream task's loss through a non-differentiable procedure.
When computing the generated sample's loss $\mathcal{L}_\text{syn}$, instead of simply averaging the losses for each generated sample as in Eq.~\eqref{eq:gradient_matching_loss}, we weight them with their generation probabilities $p_\phi(\tilde{x}_i)$ as
\begin{align}
    \mathcal{L}_\text{syn} = \sum_{i=1}^N a_i \, l_\theta(\tilde{x}_i),\\
    a_i = \frac{p_\phi(\tilde{x}_i)}{\sum_{j=1}^N p_\phi(\tilde{x}_j)}.
    %= \frac{\exp(-\mathcal{L}_\text{LM}(\tilde{x}_i))}{\sum_{j=1}^N \exp (-\mathcal{L}_\text{LM}(\tilde{x}_j))},
\end{align}
Therefore, $\mathcal{L}_\text{GM}$ can be back-propagated to $\phi$ through the differentiable pass via loss weights $a_i$, as illustrated in Figure~\ref{fig:overview}. 
Intuitively, the generator model is updated to increase its generation probabilities of synthetic samples that improve gradient similarity. 
% Moreover, this training process is also similar to reinforcement learning with policy gradient, which we discuss in Appendix~\ref{sec:appendix/connection_to_policy_gradient}.

% \paragraph{Representative teacher.}
\noindent\textbf{Representative Teacher.}
To improve DiLM, we consider enhancing the gradient teacher of real samples by using representative samples for each mini-batch of real samples instead of randomly selected ones.
Inspired by \citet{liu2023dream}, we select the representative samples with K-centers~\cite{wolf2011facility,sener2018active}, a clustering-based coreset selection method (Figure~\ref{fig:overview}a).
Specifically, we divide all the real training samples for each class into $M$ sub-clusters by using the K-means algorithm on the feature space of the learner model, and choose the center sample of each sub-cluster.
As shown in \cite{liu2023dream}, the representative samples selected by K-centers provide the proper teacher gradient by including diverse samples that cover the overall distribution for each class and eliminating samples near the decision boundaries, which have dominant gradients with large norms.
Considering coverage and robustness, we generate 10 representative sample sets by running the K-means algorithm with different random seeds at the beginning of training and use one as a mini-batch of real samples in each training step.\footnote{\citet{liu2023dream} repeatedly re-generated the K-center representative samples by conducting clustering on the feature space of the different learner model's states throughout the inner loop. However, it is very time consuming with BERT as the learner model, as in our study.}

% \paragraph{Diverse mini-batch sampling.}
\noindent\textbf{Diverse Mini-batch Sampling.}
Diversity in a mini-batch of generated samples for each step affects the sample space that the generator model explores in training. 
If the generator model only generates many samples that are similar to each other, this leads to the biased optimization of the generator model.
To address this issue, we introduce diverse mini-batch sampling of generated samples in the training process of DiLM (Figure~\ref{fig:overview}b).
Instead of generating $N$ synthetic samples for each step, the generator model generates $N\times I_\text{int}$ synthetic samples at the same time, where $I_\text{int}$ is the generation interval. The generated synthetic samples are then divided into $N$ sub-clusters with the K-means algorithm, and a mini-batch of synthetic samples for each step is constructed by randomly choosing one sample from each sub-cluster.

\subsection{Generate Synthetic Dataset}
\label{sec:sample_selection}
We obtain distilled dataset $\mathcal{D}_\text{syn}$ by generating synthetic samples with the trained generator model.
%To mitigate the influence of the bias of the generative distribution of the generator model,
To include representative samples of the model's generative distribution $p_\phi(\tilde{x})$, we use the coreset selection method again to select generated synthetic samples.
Specifically, we generate 100 times as many synthetic samples as the distilled dataset size $|\mathcal{D}_\text{syn}|$ by top-$p$ sampling with $p=0.95$, considering the diversity, and then construct $\mathcal{D}_\text{syn}$ with K-center representative samples.
This makes $\mathcal{D}_\text{syn}$ to include diverse synthetic samples %that cover the distribution of original dataset 
by removing redundant samples caused by the biased generative distribution of the model.

\section{Experimental Settings}

%\subsection{Evaluation}
\noindent\textbf{Datasets.}
We evaluated DiLM in distilling three major text classification datasets, SST-2, QQP, and MNLI-m, from the GLUE benchmark~\cite{wang-etal-2018-glue}. 
Following \citet{wang-etal-2018-glue}, we report accuracy for SST-2 and MNLI-m, and the average of accuracy and F1 score for QQP as our results.
More details about each dataset are shown in Appendix~\ref{sec:appendix/datasets}.

\noindent\textbf{Baselines.} Following previous studies on dataset distillation in the CV field, we compared the performance of DiLM with three coreset selection methods, Random, K-centers~\cite{wolf2011facility,sener2018active}, and Herding~\cite{welling2009herding}, as well as TDD~\cite{sucholutsky-2021-soft-label}, which is a recent embedding-level distillation method.
Note that TDD also trains the learnable soft-labels and learning rates for each training step together with the input word embeddings.
% We also evaluated the vanilla LM, which only learns the synthetic training data generation (Section~\ref{sec:text_generation_model}), to validate the effectiveness of the training for gradient matching, described in Section~\ref{sec:text_gtn}.
\Add{We also evaluated the vanilla LM, which skips training for gradient matching (in Section~\ref{sec:text_gtn}), to validate its effectiveness. Note that we applied K-center representative sample selection~(in Section~\ref{sec:sample_selection}) to the vanilla LM as well.}
The details of each baseline are given in Appendix~\ref{sec:appendix/baselines}.

\noindent\textbf{Evaluation.}
For evaluation, we used BERT$_\text{BASE}$ and other three pre-trained models, RoBERTa$_\text{BASE}$, BERT$_\text{LARGE}$, and XLNet$_\text{BASE}$, as learner models (see more details in Appendix~\ref{sec:appendix/learner_models}).
We trained a learner model on the distilled datasets for 200 steps by using AdamW~\cite{loshchilov2019adamw} with a learning rate of $1.0\times10^{-4}$ and a batch size of 64.\footnote{We did not follow this training protocol for TDD, since TDD optimizes learning rates as well for each step with a specific synthetic sample order.} 
For Herding and TDD, we trained the learner model on their datasets for 100 times. For other methods, we generated 20 datasets with different random seeds and trained the learner model on each of them for 5 times.
We report the average and standard deviation for these 100 models.
\Add{In the result tables, `$*$' indicates significant difference of DiLM from K-centers ($p<0.05$, Welch's t-test). Note that the standard deviations in our results inevitably become large because we trained models with few selected/generated samples from different initial model parameters. However, our evaluation procedure, which includes 100 runs, supports the reliability of our experimental results enough to discuss the effectiveness of the proposed method.}

\noindent\textbf{Implementation.}
We used the 128M parameter version of GPT-2\footnote{\url{https://huggingface.co/gpt2}} \cite{radford2019language} as the generator model of DiLM,
%For pre-training the training data generation (Section~\ref{sec:text_generation_model}), we trained the GPT-2 on each original dataset for 80000~step with AdamW with a learning rate of $1.0\times10^{-5}$ and batch size of 64.
and used BERT$_\text{BASE}$~\cite{devlin-etal-2019-bert} as the learner model, on which we calculated the gradient matching loss.
To reduce the computational costs, we calculated the gradient matching loss only for the randomly initialized last layer parameters, which tend to have dominantly larger gradient than the pre-trained parameters.
We set the number of each loop for training DiLM to $S=2000$, $T=10$, and $K=20$, and the generation interval to $I_\text{int}=200$ according to our preliminary experiments.
The mini-batch size of real and synthetic samples were respectively set to $M=200$ and $N=64$.
More details of our implementation are given in Appendix~\ref{sec:appendix/implementation_details}.

\section{Results and Discussion}
\label{sec:results}

\subsection{Performance for BERT$_\text{BASE}$}

\begin{table*}[t]
    \centering
    \begin{adjustbox}{width=\linewidth}
    \begin{tabular}{l*9c}
    \toprule
    & \multicolumn{3}{c}{\textbf{SST-2} (2 classes, 67.3k)} & \multicolumn{3}{c}{\textbf{QQP} (2 classes, 364k)} & \multicolumn{3}{c}{\textbf{MNLI-m} (3 classes, 393k)} \\
    \cmidrule(lr){2-4} \cmidrule(lr){5-7} \cmidrule(lr){8-10}
    Data/class & 5 & 10 & 20 & 5 & 10 & 20 & 5 & 10 & 20 \\
    \midrule
    Random 
    & $58.1{\pm 5.2}$ & $64.3{\pm 7.4}$ & $70.3{\pm 6.8}$  & $51.5{\pm 5.6}$ & $56.0{\pm 4.8}$ & $59.1{\pm 3.8}$ & $35.6{\pm 2.1}$ & $37.7{\pm 2.6}$ & $40.1{\pm 3.2}$  \\
    K-centers 
    & $70.8{\pm 4.1}$ & $75.9{\pm 4.7}$ & $79.8{\pm 3.5}$ & $\mathbf{60.7{\pm 3.8}}$ & $60.9{\pm 3.1}$ & $62.6{\pm 2.7}$ & $36.2{\pm 2.4}$ & $41.8{\pm 3.2}$ & $45.3{\pm 3.0}$ \\
    Herding 
    & $70.2{\pm 5.7}$ & $73.2{\pm 5.7}$ & $76.9{\pm 4.4}$ & $56.0{\pm 5.6}$ & $59.7{\pm 4.1}$ & $62.3{\pm 3.4}$ & $36.2{\pm 3.8}$ & $38.7{\pm 3.7}$ & $42.8{\pm 3.5}$ \\
    \midrule
    TDD (embed.) 
    & $89.6{\pm0.4}$ & - & -  & $81.5{\pm0.2}$ & - & - & $75.6{\pm0.2}$ & - & - \\
    TDD (text) 
    & \cellcolor{orange-red!25}$50.2{\pm1.6}$ & - & - & \cellcolor{orange-red!25}$39.6{\pm6.8}$ & - & -  & \cellcolor{orange-red!25}$33.4{\pm1.8}$ & - & -  \\
    \midrule
    Vanilla LM 
    & $65.2{\pm 6.8}$ & $71.7{\pm 6.8}$ & $77.6{\pm 4.1}$ & $56.7{\pm 4.4}$ & $59.3{\pm 3.8}$ &  $62.5{\pm 3.3}$ & $36.3{\pm 2.7}$ & $40.5{\pm 2.9}$ & $43.6{\pm 3.1}$ \\
    DiLM
    & \cellcolor{yellow-green!25}$\mathbf{72.5{\pm 5.9}^*}$ & \cellcolor{yellow-green!25}$\mathbf{76.3{\pm 4.6}}$ & \cellcolor{yellow-green!25}$\mathbf{80.3{\pm 2.8}}$ & $58.8{\pm 5.2}$ & \cellcolor{yellow-green!25}$\mathbf{62.2{\pm 3.3}^*}$ & \cellcolor{yellow-green!25}$\mathbf{64.4{\pm 2.6}^*}$ & \cellcolor{yellow-green!25}$\mathbf{39.7{\pm 2.7}^*}$ & \cellcolor{yellow-green!25}$\mathbf{44.8{\pm 3.1}^*}$ & \cellcolor{yellow-green!25}$\mathbf{48.7{\pm 2.6}^*}$ \\
    \midrule
    Full dataset & \multicolumn{3}{c}{92.7} & \multicolumn{3}{c}{89.6} & \multicolumn{3}{c}{86.7 } \\
    \bottomrule
    \end{tabular}
    \end{adjustbox}
    \caption{Performance comparison of DiLM with coreset selection methods and TDD for training the BERT$_\text{BASE}$ model. \colorbox{yellow-green!25}{Green} highlighted results indicate that DiLM outperformed the coreset selection methods. \colorbox{orange-red!25}{Red} highlighted results indicate performance degradation of distilled datasets from TDD after being converted to text. Note that we could not conduct the experiments for TDD with larger DPC settings due to GPU memory requirements.}
    \label{tab:main_results}
\end{table*}

As shown in Table~\ref{tab:main_results}, we first compared DiLM with the other baselines for training BERT$_\text{BASE}$, on which DiLM trained gradient matching. 
We evaluated them for different sizes of distilled synthetic datasets of 5/10/20 data-per-class (DPC) settings.

We first found that the vanilla LM, which was only trained for synthetic training sample generation without gradient matching, clearly underperformed the coreset selection methods. 
This indicates that, as can be expected, the quality of the generated synthetic samples becomes lower than that of real samples in the original datasets.
However, DiLM, which fine-tuned the vanilla LM with gradient matching, improved its performance and even outperformed the coreset selection methods overall. 
Note that the performance gains from K-centers indicate that DiLM generated synthetic training samples that are more effective for model training than the real samples in the original datasets.

When focusing on the difference between the three datasets, the performance gains of DiLM on QQP and MNLI-m were larger than that on SST-2.
We believe this is because QQP and MNLI-m, which are the tasks to specify the relationship between two sentences, are intuitively less likely to have real samples that represent the task than %compared with 
SST-2, which is a relatively simple negative/positive classification task.
In addition, it may also be related to the size of the original training dataset of QQP and MNLI-m, which is five times larger than that of SST-2.
% Since the generator model was trained by gradient matching with self-generated synthetic samples, pre-training with enough diversity samples in the original datasets enables the generator model to explore diverse samples in the fine-tuning phase, which results in the effective performance of DiLM.
Since the generator model was trained by gradient matching with self-generated synthetic samples, it %the generator model 
can explore broader %diverse 
sample space by pre-training with the original dataset that contains enough diversity samples, which results in the effective performance of DiLM.

For TDD, we also evaluated its distilled datasets as text data by converting them to discrete tokens that have nearest neighbor embeddings.
When directly using the distilled datasets as word embeddings, TDD achieved remarkable performance even compared with the full datasets. 
However, after converting to text, its performance catastrophically degraded even to the lower-bound performances with random prediction.
This suggests that the distilled datasets from TDD are strictly overfitted at the word embedding level and cannot be converted to text without acceptable performance degradation, which is necessary for applying them to other models.
% This is the clear limitation in contrast to DiLM.
This point is the clear advantage of DiLM, which distills synthetic datasets at the text-level.

\subsection{Cross-model Generalization}

\newcolumntype{Y}{>{\centering\arraybackslash}p{4.2em}}
\begin{table}[t]
    % \small
    \centering
    \begin{adjustbox}{width=\linewidth}
    \begin{tabular}{cl*3Y}
    \toprule
    \textbf{Dataset} & \textbf{Model}  & \textbf{Random} & \textbf{K-centers} & \textbf{DiLM} \\
    \midrule
    \multirow{4}{*}{SST-2}
    % & BERT$_\text{BASE}$ (\textbf{S}) & $70.3{\pm 6.8}$ & $79.8{\pm 3.5}$ & \cellcolor{yellow-green!25} $\mathbf{80.3{\pm 2.8}}$ \\
    & BERT$_\text{BASE}$ (\textbf{S}) & $70.3{\pm 6.8}$ & $79.8{\pm 3.5}$ & $80.3{\pm 2.8}^*$ \\
    \cmidrule{2-5}
    & RoBERTa$_\text{BASE}$ & $74.4{\pm 5.3}$ & $73.9{\pm 5.2}$ & \cellcolor{yellow-green!25} $\mathbf{78.1{\pm 3.8}^*}$ \\
    & BERT$_\text{LARGE}$ & $74.7{\pm 8.4}$ & $80.4{\pm 9.1}$ & \cellcolor{yellow-green!25} $\mathbf{83.1{\pm 6.2}^*}$ \\
    & XLNet$_\text{BASE}$ & $69.9{\pm 6.2}$ & $71.8{\pm 5.8}$ & \cellcolor{yellow-green!25} $\mathbf{77.9{\pm 4.7}^*}$ \\
    \midrule
    \multirow{4}{*}{QQP}
    % & BERT$_\text{BASE}$ (\textbf{S}) & $59.1{\pm 3.8}$ & $62.6{\pm 2.7}$ & \cellcolor{yellow-green!25} $\mathbf{64.4{\pm 2.6}}$ \\
    & BERT$_\text{BASE}$ (\textbf{S}) & $59.1{\pm 3.8}$ & $62.6{\pm 2.7}$ & $64.4{\pm 2.6^*}$ \\
    \cmidrule{2-5}
    & RoBERTa$_\text{BASE}$ & $60.1{\pm 4.0}$ & $63.9{\pm 3.2}$ & \cellcolor{yellow-green!25} $\mathbf{66.4{\pm 2.3}^*}$ \\
    & BERT$_\text{LARGE}$ & $58.8{\pm 6.9}$ & $59.0{\pm 8.9}$ & \cellcolor{yellow-green!25} $\mathbf{62.9{\pm 8.6}^*}$ \\
    & XLNet$_\text{BASE}$ & $59.1{\pm 3.5}$ & $60.9{\pm 3.0}$ & \cellcolor{yellow-green!25} $\mathbf{64.4{\pm 2.2}^*}$ \\
    \midrule
    \multirow{4}{*}{MNLI-m}
    % & BERT$_\text{BASE}$ (\textbf{S}) & $40.1{\pm 3.2}$ & $45.3{\pm 3.0}$ & \cellcolor{yellow-green!25} $\mathbf{48.7{\pm 2.6}}$ \\
    & BERT$_\text{BASE}$ (\textbf{S}) & $40.1{\pm 3.2}$ & $45.3{\pm 3.0}$ & $48.7{\pm 2.6}$ \\
    \cmidrule{2-5}
    & RoBERTa$_\text{BASE}$ & $39.6{\pm 2.5}$ & $44.5{\pm 2.6}$ & \cellcolor{yellow-green!25} $\mathbf{45.0{\pm 2.8}}$ \\
    & BERT$_\text{LARGE}$ & $40.9{\pm 4.5}$ & $48.7{\pm 4.2}$ & \cellcolor{yellow-green!25} $\mathbf{49.6{\pm 4.4}}$ \\
    & XLNet$_\text{BASE}$ & $39.0{\pm 2.0}$ & $43.5{\pm 2.7}$ & \cellcolor{yellow-green!25} $\mathbf{44.7{\pm 2.7}^*}$ \\
    \bottomrule
    \end{tabular}
    \end{adjustbox}
    \caption{Cross-model generalization performance for settings of DPC=20. (\textbf{S}) indicates the source model for gradient matching of DiLM and feature extractor for K-centers.}
    \label{tab:cross_model_generalization}
\end{table}

In contrast to the current embedding-level distillation methods, text-level synthetic datasets from DiLM can be leveraged %are available 
for training different models independent of their word embedding weights. 
To emphasize this advantage, we evaluated the distilled synthetic datasets for training three models different from BERT$_\text{BASE}$, with which the distilled synthetic datasets were obtained, 
i.e., RoBERTa$_\text{BASE}$, BERT$_\text{LARGE}$, and XLNet$_\text{BASE}$.
Table~\ref{tab:cross_model_generalization} summarizes the performances of Random, K-centers, and DiLM with DPC=20, where DiLM achieved stably good performances.\footnote{We also show the results with other DPC settings in Appendix~\ref{sec:appendix/additional_results}.} 
The results indicate that the distilled datasets from DiLM consistently performed well for training the different models, even though DiLM trained gradient matching only for the BERT$_\text{BASE}$ model's parameters. 
It is worth noting that our distilled datasets show successful generalization performance not only for training RoBERTa$_\text{BASE}$ and BERT$_\text{LARGE}$, which have the same model architecture as BERT$_\text{BASE}$, but also for training XLNet$_\text{BASE}$, which is an autoregressive model using the hidden state of the \verb|<eos>| token for classification, while BERT$_\text{BASE}$ is an autoencoding model using the hidden state of the \verb|[CLS]| token.

\begin{table}[t]
    \centering
    \begin{adjustbox}{width=\linewidth}
    \begin{tabular}{l*3c}
        \toprule
        \textbf{Models} & \textbf{Random} & \textbf{K-centers} & \textbf{DiLM} \\
        \midrule
        GPT-2-XL (1.5B) & $64.8{\pm12.0}$ & $64.8{\pm13.3}$ & \cellcolor{yellow-green!25}$\mathbf{71.1{\pm13.0}^*}$ \\
        OPT (2.7B)  & $89.3{\pm5.9}$ & $91.5{\pm3.1}$ & \cellcolor{yellow-green!25}$\mathbf{92.7{\pm1.9}^*}$ \\
        Llama~2 (7B) & $93.6{\pm2.9}$ & $94.6{\pm0.7}$ & \cellcolor{yellow-green!25}$\mathbf{95.1{\pm0.7}^*}$ \\
        \bottomrule
    \end{tabular}
    \end{adjustbox}
    \caption{Performance of distilled datasets as 5-shot prompts for in-context learning of SST-2. Each score is the average and standard deviation for 100 prompts with 20 distilled datasets and 5 random orders.}
    \label{tab:few-shot}
\end{table}

%In addition to fine-tuning models, 
We also evaluated the distilled datasets from DiLM as few-shot prompts for in-context learning of LLMs.
Table~\ref{tab:few-shot} shows the performance of Random, K-centers, and DiLM for in-context learning for SST-2 with three different sizes of LLMs, GPT-2-XL~\cite{radford2019language}, OPT~\cite{zhang2022opt}, and Llama~2~\cite{touvron2023llama2}.
Surprisingly, the distilled datasets from DiLM consistently performed well for the in-context learning, compared with Random and K-centers. %baselines.

These remarkable generalization performances across models and training processes strongly support the advantage of DiLM to distill datasets at the text-level.

% In-context learning
\subsection{Analysis and Discussion}

\noindent\textbf{Ablation Study.} Table~\ref{tab:ablation} shows the results of the ablation study for the performance improvement techniques of the representative teacher for gradient matching, the diverse mini-batch sampling of synthetic samples during training of DiLM \Add{(in Section~\ref{sec:text_gtn}), and the representative sample selection with K-centers during synthetic dataset generation (in Section~\ref{sec:sample_selection})}.
The results demonstrated that all the three techniques are consistently effective for DiLM.

\newcolumntype{Y}{>{\centering\arraybackslash}p{4.5em}}
\newcolumntype{C}{>{\centering\arraybackslash}p{2.5em}}
\begin{table}[t]
    \centering
    % \small
    \begin{adjustbox}{width=\linewidth}
    \begin{tabular}{CC>{\centering}p{3.5em}YYY}
        \toprule
        \textbf{RT} & \textbf{DMS} & \textbf{Selection} & \textbf{SST-2} & \textbf{QQP} & \textbf{MNLI-m} \\
        \midrule
        \ding{51} & \ding{51} & \ding{51} & \cellcolor{yellow-green!25}$\mathbf{72.5\pm5.9}$ & \cellcolor{yellow-green!25}$\mathbf{58.8\pm5.2}$ & \cellcolor{yellow-green!25}$\mathbf{39.7\pm2.7}$ \\
        \cmidrule{1-6}
        - & \ding{51} & \ding{51} & $70.9\pm5.9$ & $57.6\pm5.0$ & $39.5\pm2.8$ \\
        \ding{51} & - & \ding{51} & $71.3\pm5.6$ & $57.5\pm4.4$ & $38.8\pm3.0$ \\
        \ding{51} & \ding{51} & - & $65.2\pm7.0$ & $53.9\pm5.6$ & $37.9\pm3.2$ \\
        \bottomrule
    \end{tabular}
    \end{adjustbox}
    \caption{Ablation study on the performance improvement techniques of DiLM with the DPC=5 setting. RT, DMS\Add{, and Selection} indicate representative teacher, diverse mini-batch sampling, \Add{and sample selection with K-centers,} respectively.}
    \label{tab:ablation}
\end{table}

\begin{figure}[t]
    \centering
    \includegraphics[width=\linewidth]{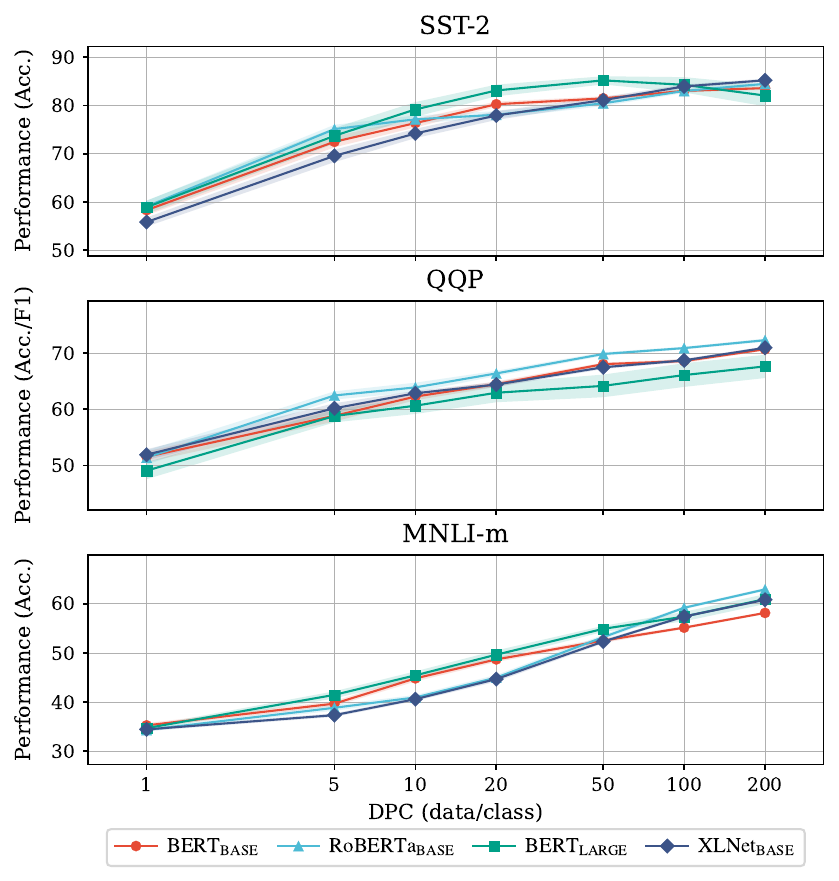}
    \caption{Performance for increasing number of synthetic samples with DPC $\in\{1,5,10,20,50,100, 200\}$. We plot the mean and 95\% confidence interval for 100 models trained on distilled datasets from DiLM.}
    \label{fig:graph_dpc}
\end{figure}

\noindent\textbf{Scaling of DPC.}
We investigated the performance of DiLM when increasing the size of synthetic datasets.
Note that DiLM does not require retraining the generator model for generating distilled synthetic datasets for different DPCs, which is also the advantage of using generative models for dataset distillation.
As shown in Figure~\ref{fig:graph_dpc}, the performance of the distilled datasets generally scaled with increasing DPC. 

\noindent\textbf{Distilled Data Examples.}
We gave examples of distilled synthetic samples for each dataset in Appendix~\ref{sec:appendix/examples}.
We found that DiLM successfully generated interpretable synthetic samples that are appropriate for the tasks of the original datasets.
Although DiLM consistently generated high quality synthetic samples for SST-2 and QQP, the repetition problem can be observed in some lengthy samples for MNLI-m.
This suggests that there is still room for performance improvements of DiLM by using a larger and more sophisticated pre-trained language model for the generator model than the small GPT-2 used in our current experiments.

\section{Conclusion}

We proposed the first text-level dataset distillation approach, called DiLM, which trains a language model to generate informative synthetic samples as text data for model-agnostic applicability and interpretability of distilled datasets.
Experimental results across various text classification datasets indicated that the distilled datasets from DiLM achieve successful performance for training various types of models beyond the source model used for distillation, even for in-context learning of LLMs.
% Although DiLM achieved remarkable performances as a text-level dataset distillation method, there are still the performance gaps from the full original datasets.
% We will improve the current text-level dataset distillation method and extend it to more difficult settings, such as text generation tasks and full-scratch training of language models, in future work.

\section*{Limitations}
The following three points are the limitations of this work.
(i) Although DiLM achieved remarkable performance as a text-level distillation method, there is still a performance gap from the full datasets.
\Add{However, the performance improvements from K-centers are large enough to demonstrate the effectiveness of DiLM, considering the fundamental difficulty of the dataset distillation problem itself in cases when synthetic data are restricted to the text-level. Moreover,}
DiLM has room for further performance improvement by employing larger and more sophisticated pre-trained language models as the generator model or using other dataset distillation objectives as an alternative to the gradient matching.
(ii) In our experiments, we applied DiLM to distill only text classification task datasets. 
% We will explore the text dataset distillation for more difficult settings, such as text generation tasks and full-scratch training of language models, in future work.
DiLM can be applied to text generation tasks as well by just considering the entire original training dataset as the data for a single label. 
In future work, we should explore the application of DiLM for more difficult settings, such as the text generation tasks and full-scratch training of language models. 
(iii) While privacy preservation of the original training datasets is one of the applications of dataset distillation, it is difficult to apply DiLM to the privacy preservation because the distilled synthetic datasets from DiLM may include real samples from the original dataset due to the training data memorization of the language model.
However, we believe that the advantage of DiLM to generate distilled synthetic datasets at the text-level, enabling the training of models independent of word embedding weights, is more valuable than the application to the privacy preservation in terms of practical applications.

\bibliography{anthology,custom}

\begin{thebibliography}{45}
\expandafter\ifx\csname natexlab\endcsname\relax\def\natexlab#1{#1}\fi

\bibitem[{Bohdal et~al.(2020)Bohdal, Yang, and Hospedales}]{bohdal-2020-flexible-dataset}
Ondrej Bohdal, Yongxin Yang, and Timothy~M. Hospedales. 2020.
\newblock \href {http://arxiv.org/abs/2006.08572} {Flexible dataset distillation: Learn labels instead of images}.
\newblock \emph{CoRR}, abs/2006.08572.

\bibitem[{Brown et~al.(2020)Brown, Mann, Ryder, Subbiah, Kaplan, Dhariwal, Neelakantan, Shyam, Sastry, Askell, Agarwal, Herbert-Voss, Krueger, Henighan, Child, Ramesh, Ziegler, Wu, Winter, Hesse, Chen, Sigler, Litwin, Gray, Chess, Clark, Berner, McCandlish, Radford, Sutskever, and Amodei}]{brown2020gpt3}
Tom Brown, Benjamin Mann, Nick Ryder, Melanie Subbiah, Jared~D Kaplan, Prafulla Dhariwal, Arvind Neelakantan, Pranav Shyam, Girish Sastry, Amanda Askell, Sandhini Agarwal, Ariel Herbert-Voss, Gretchen Krueger, Tom Henighan, Rewon Child, Aditya Ramesh, Daniel Ziegler, Jeffrey Wu, Clemens Winter, Chris Hesse, Mark Chen, Eric Sigler, Mateusz Litwin, Scott Gray, Benjamin Chess, Jack Clark, Christopher Berner, Sam McCandlish, Alec Radford, Ilya Sutskever, and Dario Amodei. 2020.
\newblock \href {https://proceedings.neurips.cc/paper_files/paper/2020/file/1457c0d6bfcb4967418bfb8ac142f64a-Paper.pdf} {Language models are few-shot learners}.
\newblock In \emph{Advances in Neural Information Processing Systems}, volume~33, pages 1877--1901. Curran Associates, Inc.

\bibitem[{Cazenavette et~al.(2022)Cazenavette, Wang, Torralba, Efros, and Zhu}]{Cazenavette_2022_CVPR}
George Cazenavette, Tongzhou Wang, Antonio Torralba, Alexei~A. Efros, and Jun{-}Yan Zhu. 2022.
\newblock \href {https://doi.org/10.1109/CVPRW56347.2022.00521} {Dataset distillation by matching training trajectories}.
\newblock In \emph{{IEEE/CVF} Conference on Computer Vision and Pattern Recognition Workshops, {CVPR} Workshops 2022, New Orleans, LA, USA, June 19-20, 2022}, pages 4749--4758. {IEEE}.

\bibitem[{Cazenavette et~al.(2023)Cazenavette, Wang, Torralba, Efros, and Zhu}]{cazenavette2023generative}
George Cazenavette, Tongzhou Wang, Antonio Torralba, Alexei~A. Efros, and Jun{-}Yan Zhu. 2023.
\newblock \href {https://doi.org/10.1109/CVPR52729.2023.00364} {Generalizing dataset distillation via deep generative prior}.
\newblock In \emph{{IEEE/CVF} Conference on Computer Vision and Pattern Recognition, {CVPR} 2023, Vancouver, BC, Canada, June 17-24, 2023}, pages 3739--3748. {IEEE}.

\bibitem[{Chen et~al.(2022)Chen, Kerkouche, and Fritz}]{chen-2022-private-set}
Dingfan Chen, Raouf Kerkouche, and Mario Fritz. 2022.
\newblock \href {https://proceedings.neurips.cc/paper_files/paper/2022/file/5e1a87dbb7e954b8d9d6c91f6db771eb-Paper-Conference.pdf} {Private set generation with discriminative information}.
\newblock In \emph{Advances in Neural Information Processing Systems}, volume~35, pages 14678--14690. Curran Associates, Inc.

\bibitem[{Cui et~al.(2023)Cui, Wang, Si, and Hsieh}]{pmlr-v202-cui23e}
Justin Cui, Ruochen Wang, Si~Si, and Cho-Jui Hsieh. 2023.
\newblock \href {https://proceedings.mlr.press/v202/cui23e.html} {Scaling up dataset distillation to {I}mage{N}et-1{K} with constant memory}.
\newblock In \emph{Proceedings of the 40th International Conference on Machine Learning}, volume 202 of \emph{Proceedings of Machine Learning Research}, pages 6565--6590. PMLR.

\bibitem[{Devlin et~al.(2019)Devlin, Chang, Lee, and Toutanova}]{devlin-etal-2019-bert}
Jacob Devlin, Ming-Wei Chang, Kenton Lee, and Kristina Toutanova. 2019.
\newblock \href {https://doi.org/10.18653/v1/N19-1423} {{BERT}: Pre-training of deep bidirectional transformers for language understanding}.
\newblock In \emph{Proceedings of the 2019 Conference of the North {A}merican Chapter of the Association for Computational Linguistics: Human Language Technologies, Volume 1 (Long and Short Papers)}, pages 4171--4186, Minneapolis, Minnesota. Association for Computational Linguistics.

\bibitem[{Dong et~al.(2022)Dong, Zhao, and Lyu}]{pmlr-v162-dong22c}
Tian Dong, Bo~Zhao, and Lingjuan Lyu. 2022.
\newblock \href {https://proceedings.mlr.press/v162/dong22c.html} {Privacy for free: How does dataset condensation help privacy?}
\newblock In \emph{Proceedings of the 39th International Conference on Machine Learning}, volume 162 of \emph{Proceedings of Machine Learning Research}, pages 5378--5396. PMLR.

\bibitem[{Geng et~al.(2023)Geng, Chen, Wang, Woisetschlaeger, Schimmler, Mayer, Zhao, and Rong}]{Geng2023survey}
Jiahui Geng, Zongxiong Chen, Yuandou Wang, Herbert Woisetschlaeger, Sonja Schimmler, Ruben Mayer, Zhiming Zhao, and Chunming Rong. 2023.
\newblock \href {https://doi.org/10.24963/ijcai.2023/741} {A survey on dataset distillation: Approaches, applications and future directions}.
\newblock In \emph{Proceedings of the Thirty-Second International Joint Conference on Artificial Intelligence, {IJCAI-23}}, pages 6610--6618. International Joint Conferences on Artificial Intelligence Organization.
\newblock Survey Track.

\bibitem[{Goodfellow et~al.(2014)Goodfellow, Pouget-Abadie, Mirza, Xu, Warde-Farley, Ozair, Courville, and Bengio}]{ian2014gan}
Ian Goodfellow, Jean Pouget-Abadie, Mehdi Mirza, Bing Xu, David Warde-Farley, Sherjil Ozair, Aaron Courville, and Yoshua Bengio. 2014.
\newblock \href {https://proceedings.neurips.cc/paper_files/paper/2014/file/5ca3e9b122f61f8f06494c97b1afccf3-Paper.pdf} {Generative adversarial nets}.
\newblock In \emph{Advances in Neural Information Processing Systems}, volume~27. Curran Associates, Inc.

\bibitem[{Hiraoka et~al.(2020)Hiraoka, Takase, Uchiumi, Keyaki, and Okazaki}]{hiraoka-etal-2020-optimizing}
Tatsuya Hiraoka, Sho Takase, Kei Uchiumi, Atsushi Keyaki, and Naoaki Okazaki. 2020.
\newblock \href {https://doi.org/10.18653/v1/2020.findings-emnlp.120} {Optimizing word segmentation for downstream task}.
\newblock In \emph{Findings of the Association for Computational Linguistics: EMNLP 2020}, pages 1341--1351, Online. Association for Computational Linguistics.

\bibitem[{Li and Li(2021)}]{li-2021-data-distillation}
Yongqi Li and Wenjie Li. 2021.
\newblock \href {http://arxiv.org/abs/2104.08448} {Data distillation for text classification}.
\newblock \emph{CoRR}, abs/2104.08448.

\bibitem[{Liu et~al.(2023)Liu, Gu, Wang, Zhu, Jiang, and You}]{liu2023dream}
Yanqing Liu, Jianyang Gu, Kai Wang, Zheng Zhu, Wei Jiang, and Yang You. 2023.
\newblock Dream: Efficient dataset distillation by representative matching.
\newblock In \emph{Proceedings of the IEEE/CVF International Conference on Computer Vision (ICCV)}, pages 17314--17324.

\bibitem[{Liu et~al.(2019)Liu, Ott, Goyal, Du, Joshi, Chen, Levy, Lewis, Zettlemoyer, and Stoyanov}]{liu-2019-roberta}
Yinhan Liu, Myle Ott, Naman Goyal, Jingfei Du, Mandar Joshi, Danqi Chen, Omer Levy, Mike Lewis, Luke Zettlemoyer, and Veselin Stoyanov. 2019.
\newblock \href {http://arxiv.org/abs/1907.11692} {Roberta: {A} robustly optimized {BERT} pretraining approach}.
\newblock \emph{CoRR}, abs/1907.11692.

\bibitem[{Loshchilov and Hutter(2019)}]{loshchilov2019adamw}
Ilya Loshchilov and Frank Hutter. 2019.
\newblock \href {https://openreview.net/forum?id=Bkg6RiCqY7} {Decoupled weight decay regularization}.
\newblock In \emph{7th International Conference on Learning Representations, {ICLR} 2019, New Orleans, LA, USA, May 6-9, 2019}. OpenReview.net.

\bibitem[{Maclaurin et~al.(2015)Maclaurin, Duvenaud, and Adams}]{maclaurin2015gradient}
Dougal Maclaurin, David Duvenaud, and Ryan~P. Adams. 2015.
\newblock \href {http://proceedings.mlr.press/v37/maclaurin15.html} {Gradient-based hyperparameter optimization through reversible learning}.
\newblock In \emph{Proceedings of the 32nd International Conference on Machine Learning, {ICML} 2015, Lille, France, 6-11 July 2015}, volume~37 of \emph{{JMLR} Workshop and Conference Proceedings}, pages 2113--2122. JMLR.org.

\bibitem[{Maekawa et~al.(2023)Maekawa, Kobayashi, Funakoshi, and Okumura}]{maekawa-etal-2023-dataset}
Aru Maekawa, Naoki Kobayashi, Kotaro Funakoshi, and Manabu Okumura. 2023.
\newblock \href {https://doi.org/10.18653/v1/2023.acl-short.12} {Dataset distillation with attention labels for fine-tuning {BERT}}.
\newblock In \emph{Proceedings of the 61st Annual Meeting of the Association for Computational Linguistics (Volume 2: Short Papers)}, pages 119--127, Toronto, Canada. Association for Computational Linguistics.

\bibitem[{Medvedev and D'yakonov(2021)}]{medvedev2022learning}
Dmitry Medvedev and Alexander D'yakonov. 2021.
\newblock \href {https://doi.org/10.1007/978-3-031-15168-2\_12} {Learning to generate synthetic training data using gradient matching and implicit differentiation}.
\newblock In \emph{Recent Trends in Analysis of Images, Social Networks and Texts - 10th International Conference, {AIST} 2021, Tbilisi, Georgia, December 16-18, 2021, Revised Supplementary Proceedings}, volume 1573 of \emph{Communications in Computer and Information Science}, pages 138--150. Springer.

\bibitem[{Pennington et~al.(2014)Pennington, Socher, and Manning}]{pennington-etal-2014-glove}
Jeffrey Pennington, Richard Socher, and Christopher Manning. 2014.
\newblock \href {https://doi.org/10.3115/v1/D14-1162} {{G}lo{V}e: Global vectors for word representation}.
\newblock In \emph{Proceedings of the 2014 Conference on Empirical Methods in Natural Language Processing ({EMNLP})}, pages 1532--1543, Doha, Qatar. Association for Computational Linguistics.

\bibitem[{Radford et~al.(2019)Radford, Wu, Child, Luan, Amodei, Sutskever et~al.}]{radford2019language}
Alec Radford, Jeffrey Wu, Rewon Child, David Luan, Dario Amodei, Ilya Sutskever, et~al. 2019.
\newblock Language models are unsupervised multitask learners.
\newblock \emph{OpenAI blog}, 1(8):9.

\bibitem[{Sahni and Patel(2023)}]{sahni2023multilingual}
Shivam Sahni and Harsh~M. Patel. 2023.
\newblock \href {https://doi.org/10.48550/arXiv.2308.04982} {Exploring multilingual text data distillation}.
\newblock \emph{CoRR}, abs/2308.04982.

\bibitem[{Sangermano et~al.(2022)Sangermano, Carta, Cossu, and Bacciu}]{sangermano-2022-sample-condensation}
Mattia Sangermano, Antonio Carta, Andrea Cossu, and Davide Bacciu. 2022.
\newblock \href {https://doi.org/10.1109/IJCNN55064.2022.9892299} {Sample condensation in online continual learning}.
\newblock In \emph{International Joint Conference on Neural Networks, {IJCNN} 2022, Padua, Italy, July 18-23, 2022}, pages 1--8. {IEEE}.

\bibitem[{Sener and Savarese(2018)}]{sener2018active}
Ozan Sener and Silvio Savarese. 2018.
\newblock \href {https://openreview.net/forum?id=H1aIuk-RW} {Active learning for convolutional neural networks: {A} core-set approach}.
\newblock In \emph{6th International Conference on Learning Representations, {ICLR} 2018, Vancouver, BC, Canada, April 30 - May 3, 2018, Conference Track Proceedings}. OpenReview.net.

\bibitem[{Such et~al.(2020)Such, Rawal, Lehman, Stanley, and Clune}]{pmlr-v119-such20a}
Felipe~Petroski Such, Aditya Rawal, Joel Lehman, Kenneth Stanley, and Jeffrey Clune. 2020.
\newblock \href {https://proceedings.mlr.press/v119/such20a.html} {Generative teaching networks: Accelerating neural architecture search by learning to generate synthetic training data}.
\newblock In \emph{Proceedings of the 37th International Conference on Machine Learning}, volume 119 of \emph{Proceedings of Machine Learning Research}, pages 9206--9216. PMLR.

\bibitem[{Sucholutsky and Schonlau(2021)}]{sucholutsky-2021-soft-label}
Ilia Sucholutsky and Matthias Schonlau. 2021.
\newblock \href {https://doi.org/10.1109/IJCNN52387.2021.9533769} {Soft-label dataset distillation and text dataset distillation}.
\newblock In \emph{2021 International Joint Conference on Neural Networks (IJCNN)}, pages 1--8.

\bibitem[{Touvron et~al.(2023)Touvron, Martin, Stone, Albert, Almahairi, Babaei, Bashlykov, Batra, Bhargava, Bhosale, Bikel, Blecher, Canton{-}Ferrer, Chen, Cucurull, Esiobu, Fernandes, Fu, Fu, Fuller, Gao, Goswami, Goyal, Hartshorn, Hosseini, Hou, Inan, Kardas, Kerkez, Khabsa, Kloumann, Korenev, Koura, Lachaux, Lavril, Lee, Liskovich, Lu, Mao, Martinet, Mihaylov, Mishra, Molybog, Nie, Poulton, Reizenstein, Rungta, Saladi, Schelten, Silva, Smith, Subramanian, Tan, Tang, Taylor, Williams, Kuan, Xu, Yan, Zarov, Zhang, Fan, Kambadur, Narang, Rodriguez, Stojnic, Edunov, and Scialom}]{touvron2023llama2}
Hugo Touvron, Louis Martin, Kevin Stone, Peter Albert, Amjad Almahairi, Yasmine Babaei, Nikolay Bashlykov, Soumya Batra, Prajjwal Bhargava, Shruti Bhosale, Dan Bikel, Lukas Blecher, Cristian Canton{-}Ferrer, Moya Chen, Guillem Cucurull, David Esiobu, Jude Fernandes, Jeremy Fu, Wenyin Fu, Brian Fuller, Cynthia Gao, Vedanuj Goswami, Naman Goyal, Anthony Hartshorn, Saghar Hosseini, Rui Hou, Hakan Inan, Marcin Kardas, Viktor Kerkez, Madian Khabsa, Isabel Kloumann, Artem Korenev, Punit~Singh Koura, Marie{-}Anne Lachaux, Thibaut Lavril, Jenya Lee, Diana Liskovich, Yinghai Lu, Yuning Mao, Xavier Martinet, Todor Mihaylov, Pushkar Mishra, Igor Molybog, Yixin Nie, Andrew Poulton, Jeremy Reizenstein, Rashi Rungta, Kalyan Saladi, Alan Schelten, Ruan Silva, Eric~Michael Smith, Ranjan Subramanian, Xiaoqing~Ellen Tan, Binh Tang, Ross Taylor, Adina Williams, Jian~Xiang Kuan, Puxin Xu, Zheng Yan, Iliyan Zarov, Yuchen Zhang, Angela Fan, Melanie Kambadur, Sharan Narang, Aur{\'{e}}lien Rodriguez, Robert Stojnic, Sergey Edunov,
  and Thomas Scialom. 2023.
\newblock \href {https://doi.org/10.48550/ARXIV.2307.09288} {Llama 2: Open foundation and fine-tuned chat models}.
\newblock \emph{CoRR}, abs/2307.09288.

\bibitem[{Wang et~al.(2018{\natexlab{a}})Wang, Singh, Michael, Hill, Levy, and Bowman}]{wang-etal-2018-glue}
Alex Wang, Amanpreet Singh, Julian Michael, Felix Hill, Omer Levy, and Samuel Bowman. 2018{\natexlab{a}}.
\newblock \href {https://doi.org/10.18653/v1/W18-5446} {{GLUE}: A multi-task benchmark and analysis platform for natural language understanding}.
\newblock In \emph{Proceedings of the 2018 {EMNLP} Workshop {B}lackbox{NLP}: Analyzing and Interpreting Neural Networks for {NLP}}, pages 353--355, Brussels, Belgium. Association for Computational Linguistics.

\bibitem[{Wang et~al.(2023)Wang, Gu, Zhou, Zhu, Jiang, and You}]{wang2023dim}
Kai Wang, Jianyang Gu, Daquan Zhou, Zheng Zhu, Wei Jiang, and Yang You. 2023.
\newblock \href {https://doi.org/10.48550/ARXIV.2303.04707} {Dim: Distilling dataset into generative model}.
\newblock \emph{CoRR}, abs/2303.04707.

\bibitem[{Wang et~al.(2022)Wang, Zhao, Peng, Zhu, Yang, Wang, Huang, Bilen, Wang, and You}]{wang2022cafe}
Kai Wang, Bo~Zhao, Xiangyu Peng, Zheng Zhu, Shuo Yang, Shuo Wang, Guan Huang, Hakan Bilen, Xinchao Wang, and Yang You. 2022.
\newblock \href {https://doi.org/10.1109/CVPR52688.2022.01188} {Cafe: Learning to condense dataset by aligning features}.
\newblock In \emph{2022 IEEE/CVF Conference on Computer Vision and Pattern Recognition (CVPR)}, pages 12186--12195.

\bibitem[{Wang et~al.(2018{\natexlab{b}})Wang, Zhu, Torralba, and Efros}]{wang2018dataset}
Tongzhou Wang, Jun{-}Yan Zhu, Antonio Torralba, and Alexei~A. Efros. 2018{\natexlab{b}}.
\newblock \href {http://arxiv.org/abs/1811.10959} {Dataset distillation}.
\newblock \emph{CoRR}, abs/1811.10959.

\bibitem[{Welling(2009)}]{welling2009herding}
Max Welling. 2009.
\newblock \href {https://doi.org/10.1145/1553374.1553517} {Herding dynamical weights to learn}.
\newblock In \emph{Proceedings of the 26th Annual International Conference on Machine Learning}, ICML '09, page 1121–1128, New York, NY, USA. Association for Computing Machinery.

\bibitem[{Wiewel and Yang(2021)}]{wiewel-2021-condensed-conposite-memory}
Felix Wiewel and Bin Yang. 2021.
\newblock \href {https://doi.org/10.1109/IJCNN52387.2021.9533491} {Condensed composite memory continual learning}.
\newblock In \emph{International Joint Conference on Neural Networks, {IJCNN} 2021, Shenzhen, China, July 18-22, 2021}, pages 1--8. {IEEE}.

\bibitem[{Wolf(2011)}]{wolf2011facility}
Gert~W. Wolf. 2011.
\newblock \href {https://doi.org/10.1080/13658816.2010.528422} {Facility location: concepts, models, algorithms and case studies. series: Contributions to management science}.
\newblock \emph{Int. J. Geogr. Inf. Sci.}, 25(2):331--333.

\bibitem[{Xiong et~al.(2023)Xiong, Wang, Cheng, Yu, and Hsieh}]{xiong2023feddm}
Yuanhao Xiong, Ruochen Wang, Minhao Cheng, Felix Yu, and Cho{-}Jui Hsieh. 2023.
\newblock \href {https://doi.org/10.1109/CVPR52729.2023.01566} {Feddm: Iterative distribution matching for communication-efficient federated learning}.
\newblock In \emph{{IEEE/CVF} Conference on Computer Vision and Pattern Recognition, {CVPR} 2023, Vancouver, BC, Canada, June 17-24, 2023}, pages 16323--16332. {IEEE}.

\bibitem[{Yang et~al.(2019)Yang, Dai, Yang, Carbonell, Salakhutdinov, and Le}]{yang-etal-2019-xlnet}
Zhilin Yang, Zihang Dai, Yiming Yang, Jaime~G. Carbonell, Ruslan Salakhutdinov, and Quoc~V. Le. 2019.
\newblock \href {https://proceedings.neurips.cc/paper/2019/hash/dc6a7e655d7e5840e66733e9ee67cc69-Abstract.html} {Xlnet: Generalized autoregressive pretraining for language understanding}.
\newblock In \emph{Advances in Neural Information Processing Systems 32: Annual Conference on Neural Information Processing Systems 2019, NeurIPS 2019, December 8-14, 2019, Vancouver, BC, Canada}, pages 5754--5764.

\bibitem[{Yu et~al.(2017)Yu, Zhang, Wang, and Yu}]{yu-etal-2017-seqgan}
Lantao Yu, Weinan Zhang, Jun Wang, and Yong Yu. 2017.
\newblock \href {https://doi.org/10.1609/AAAI.V31I1.10804} {Seqgan: Sequence generative adversarial nets with policy gradient}.
\newblock In \emph{Proceedings of the Thirty-First {AAAI} Conference on Artificial Intelligence, February 4-9, 2017, San Francisco, California, {USA}}, pages 2852--2858. {AAAI} Press.

\bibitem[{Yu et~al.(2023)Yu, Liu, and Wang}]{yu2023review}
Ruonan Yu, Songhua Liu, and Xinchao Wang. 2023.
\newblock \href {https://doi.org/10.48550/ARXIV.2301.07014} {Dataset distillation: {A} comprehensive review}.
\newblock \emph{CoRR}, abs/2301.07014.

\bibitem[{Zhang et~al.(2022{\natexlab{a}})Zhang, Chen, Li, Lyu, Wu, Ding, Shen, and Wu}]{zhang2022dense}
Jie Zhang, Chen Chen, Bo~Li, Lingjuan Lyu, Shuang Wu, Shouhong Ding, Chunhua Shen, and Chao Wu. 2022{\natexlab{a}}.
\newblock \href {https://proceedings.neurips.cc/paper_files/paper/2022/file/868f2266086530b2c71006ea1908b14a-Paper-Conference.pdf} {Dense: Data-free one-shot federated learning}.
\newblock In \emph{Advances in Neural Information Processing Systems}, volume~35, pages 21414--21428. Curran Associates, Inc.

\bibitem[{Zhang et~al.(2022{\natexlab{b}})Zhang, Roller, Goyal, Artetxe, Chen, Chen, Dewan, Diab, Li, Lin, Mihaylov, Ott, Shleifer, Shuster, Simig, Koura, Sridhar, Wang, and Zettlemoyer}]{zhang2022opt}
Susan Zhang, Stephen Roller, Naman Goyal, Mikel Artetxe, Moya Chen, Shuohui Chen, Christopher Dewan, Mona~T. Diab, Xian Li, Xi~Victoria Lin, Todor Mihaylov, Myle Ott, Sam Shleifer, Kurt Shuster, Daniel Simig, Punit~Singh Koura, Anjali Sridhar, Tianlu Wang, and Luke Zettlemoyer. 2022{\natexlab{b}}.
\newblock \href {https://doi.org/10.48550/ARXIV.2205.01068} {{OPT:} open pre-trained transformer language models}.
\newblock \emph{CoRR}, abs/2205.01068.

\bibitem[{Zhang et~al.(2016)Zhang, Gan, and Carin}]{zhang-etal-2016-softargmax}
Y~Zhang, Z~Gan, and L~Carin. 2016.
\newblock Generating text via adversarial training.
\newblock In \emph{NIPS workshop on Adversarial Training}. academia. edu.

\bibitem[{Zhang et~al.(2017)Zhang, Gan, Fan, Chen, Henao, Shen, and Carin}]{zhang-etal-2017-textgan}
Yizhe Zhang, Zhe Gan, Kai Fan, Zhi Chen, Ricardo Henao, Dinghan Shen, and Lawrence Carin. 2017.
\newblock \href {https://proceedings.mlr.press/v70/zhang17b.html} {Adversarial feature matching for text generation}.
\newblock In \emph{Proceedings of the 34th International Conference on Machine Learning}, volume~70 of \emph{Proceedings of Machine Learning Research}, pages 4006--4015. PMLR.

\bibitem[{Zhao and Bilen(2021)}]{pmlr-v139-zhao21a}
Bo~Zhao and Hakan Bilen. 2021.
\newblock \href {https://proceedings.mlr.press/v139/zhao21a.html} {Dataset condensation with differentiable siamese augmentation}.
\newblock In \emph{Proceedings of the 38th International Conference on Machine Learning}, volume 139 of \emph{Proceedings of Machine Learning Research}, pages 12674--12685. PMLR.

\bibitem[{Zhao and Bilen(2022)}]{zhao2022itgan}
Bo~Zhao and Hakan Bilen. 2022.
\newblock \href {https://doi.org/10.48550/ARXIV.2204.07513} {Synthesizing informative training samples with {GAN}}.
\newblock \emph{CoRR}, abs/2204.07513.

\bibitem[{Zhao and Bilen(2023)}]{zhao2023distribution}
Bo~Zhao and Hakan Bilen. 2023.
\newblock \href {https://doi.org/10.1109/WACV56688.2023.00645} {Dataset condensation with distribution matching}.
\newblock In \emph{{IEEE/CVF} Winter Conference on Applications of Computer Vision, {WACV} 2023, Waikoloa, HI, USA, January 2-7, 2023}, pages 6503--6512. {IEEE}.

\bibitem[{Zhao et~al.(2021)Zhao, Mopuri, and Bilen}]{zhao2021dataset}
Bo~Zhao, Konda~Reddy Mopuri, and Hakan Bilen. 2021.
\newblock \href {https://openreview.net/forum?id=mSAKhLYLSsl} {Dataset condensation with gradient matching}.
\newblock In \emph{9th International Conference on Learning Representations, {ICLR} 2021, Virtual Event, Austria, May 3-7, 2021}. OpenReview.net.

\end{thebibliography}

\clearpage

\appendix

% \section{Connection to Policy Gradient}
% \label{sec:appendix/connection_to_policy_gradient}

% \section{Application to Text Generation Tasks}

% Although we targeted at text classification tasks in our experiments, DiLM can be applied to text generation tasks as well by simply considering the entire original training dataset as the same single label data.
% Specifically, the generator model is trained with the common beginning-of-sentence tokens \verb|<bos>| for all samples as
% \begin{quote}
%     \verb|<bos>| \textit{sentence} \verb|<eos>|.
% \end{quote}
% For sequence-to-sequence generation tasks, we can use the separate token \verb|<sep>| to split source and target sentences as 
% \begin{quote}
%     \verb|<bos>| \textit{source sentence} \verb|<sep>| \textit{target sentence} \verb|<eos>|.
% \end{quote}
% Gradient matching loss is calculated with K-center representative real samples of the whole original dataset and synthetic samples generated from \verb|<bos>|.

\section{Background Details of DiLM}
\label{sec:appendix/background}

% Rebuttal comment:
% the standard approaches you mentioned cannot be applied to this case, while, in fact, we also tried to consider it at first. As for the policy gradient, while we can apply it with the sample-level gradient similarity as the reward function, the gradients of synthetic data in the true objective of gradient matching should be calculated as a set of generated samples, not a single sample, and calculating per-sample-gradient is computationally inefficient. In addition, our method can also be formulated as the policy gradient, while the reward function is somewhat too complicated, that is one of the reasons why we didn’t mention it in the paper. The soft-argmax and the same type of approaches are also not applicable, due to the difference of the vocabulary between generator and learner models.

\Add{%
In this section, we provide the detailed background of the techniques that we introduced in Section~\ref{sec:text_gtn} to enable back-propagation computing by bypassing the non-differentiable discrete text via loss weighting according to the generation probabilities. 
}

\Add{%
As we described in Section~\ref{sec:text_gtn}, although there existed two standard approaches to the non-differentiable problem in the discrete text generation, that is, soft-argmax~\cite{zhang-etal-2016-softargmax} and policy gradient~\cite{yu-etal-2017-seqgan}, both of them cannot be applied in training DiLM.
For soft-argmax and the same type of approaches, it is necessary that the vocabulary of the generator model be the same as that of the learner model, which receives the text generated by the generator model as an input.
However, it is not true in the case of DiLM, where we used GPT-2 for the generator model and BERT$_\text{BASE}$ for the learner model.
}

\Add{%
As for policy gradient, we can apply it with sample-level gradient similarity as the reward function.
However, the gradients for synthetic data for gradient matching loss should be calculated as an average for samples in a mini-batch, not for a single sample. 
Moreover, calculating per-sample gradients is computationally inefficient.
These are the reasons why we did not use the policy gradient with the per-sample gradient similarity.
}

\Add{%
However, the basic idea of our approach, which aims to update the generator model to increase its generation probabilities for synthetic samples that improve gradient similarity, is essentially the same as the policy gradient.
In addition, it is worth noting that our approach can also be formulated as the policy gradient manner.
Letting $g_\theta(\tilde{x})=\frac{\partial l_\theta(\tilde{x})}{\partial\theta}$ and $r(\cdot)$ be the reward function, the gradient of the generator parameters $\phi$ is represented as
\begin{align*}
    & \nabla_\phi\mathcal{L}_\phi = \sum_{n=1}^N \nabla_\phi l_\phi (\tilde{x}_n)\cdot r(\tilde{x}_n)  \quad \text{where}\\
    & r(\tilde{x}_n) = a_n\left\{
    g_\theta(\tilde{x}_n) - g_\theta(\tilde{X})\right\}^T
    \left( -\frac{\partial \mathcal{L}_\mathrm{GM}}{\partial g_\theta(\tilde{X})} \right).
\end{align*}
}

\section{Datasets}
\label{sec:appendix/datasets}

We used three text classification datasets in the GLUE benchmark~\cite{wang-etal-2018-glue} from huggingface datasets.\footnote{\url{https://huggingface.co/datasets/glue}}
SST-2 is a banally sentiment classification (negative/positive) task for movie review sentences.
QQP is a task to identify whether a question pair is semantically equivalent or not.
MNLI-m is a natural language inference task to predict a premise sentence entails or contradicts a hypothesis sentence or neither (neutral). 
We reported the evaluation results on the validation set in Section~\ref{sec:results}, since the test set is not publicly available.
For MNLI-m, we used the matched-domain validation set for evaluation.
We summarize the statistics of each dataset in Table~\ref{tab:datasets}. 

\begin{table}[ht!]
    \centering
    \begin{adjustbox}{width=\linewidth}
    \begin{tabular}{ccrrr}
    \toprule
    \textbf{Dataset} & \textbf{Metric} & \textbf{\#Train} & \textbf{\#Dev} & \textbf{\#Class}\\
    \midrule
    SST-2 & accuracy & 67k & 872 & 2 \\
    QQP  & accuracy/F1 & 364k & 40k & 2 \\
    MNLI-m & accuracy & 393k & 9.8k & 3 \\
    \bottomrule
    \end{tabular}
    \end{adjustbox}
    \caption{Summary of statistics of evaluation datasets}
    \label{tab:datasets}
\end{table}

\section{Baselines Details}
\label{sec:appendix/baselines}

In this section, we explain the details of the baseline methods used in our experiments.

\subsection{Coreset Selection}

\noindent\textbf{Random} is the simplest baseline, which randomly selects real samples from the original training dataset.

\noindent\textbf{K-centers}~\cite{wolf2011facility,sener2018active} is a standard coreset selection method that selects the center samples of sub-clusters as a coreset, which eliminates redundant samples and covers the distribution of the original dataset.

\noindent\textbf{Herding}~\cite{welling2009herding} is also a standard coreset selection method that greedily selects real samples to match their mean embedding with that of the original dataset.

For K-centers and Herding, we used the last hidden state of the \verb|[CLS]| token in the BERT$_\text{BASE}$ model as a feature of each training sample. 

\subsection{Embedding-level Dataset Distillation}

\noindent\textbf{TDD}\footnote{We used the implementation by \citet{maekawa-etal-2023-dataset}, because it also employs BERT as the learner model.}~\cite{sucholutsky-2021-soft-label} is the current embedding level text dataset distillation method. 
TDD also optimizes learnable soft-labels and learning rates together with input word embeddings by the original meta-learning approach~\cite{wang2018dataset}.
Following the best performing settings in \citet{maekawa-etal-2023-dataset}, which applied this approach to the BERT model, we used one synthetic sample per class as a mini-batch of a single gradient step and fixed the order of synthetic samples, which means the learner model is trained with 5 gradient steps in the experiments in Section~\ref{sec:results} with DPC=5. 
Similar to DiLM, TDD also used BERT$_\text{BASE}$ as the learner model for distillation.

\section{Learner Models}
\label{sec:appendix/learner_models}

\noindent\textbf{BERT$_\text{BASE}$}\footnote{\url{https://huggingface.co/bert-base-uncased}}~\cite{devlin-etal-2019-bert} was used as the source model for training for dataset distillation and the feature extractor of the coreset selection methods. Following the fine-tuning settings in \citet{devlin-etal-2019-bert}, we used a randomly initialized linear layer on the top of the last hidden state of the \verb|[CLS]| token.

\noindent\textbf{RoBERTa$_\text{BASE}$}\footnote{\url{https://huggingface.co/roberta-base}} is a BERT derivative model proposed by \citet{liu-2019-roberta}. This model has the same size and architecture as BERT$_\text{BASE}$, but has different parameters pre-trained with the masked language modeling (MLM) task, without the next sentence prediction (NSP) task, on a larger corpus than the BERT models.

\noindent\textbf{BERT$_\text{LARGE}$}\footnote{\url{https://huggingface.co/bert-large-uncased}} is the 24 layer, 340M parameter version of BERT, while BERT$_\text{BASE}$ has 12 layers and 110M parameters.

\noindent\textbf{XLNet$_\text{BASE}$}\footnote{\url{https://huggingface.co/xlnet-base-cased}} is an autoregressive model in contrast to BERT and RoBERTa. 
Following \cite{yang-etal-2019-xlnet}, we used a randomly initialized linear layer on the top of the last hidden state of the \verb|<eos>| token, which involves entire tokens in the sequence.

\section{Implementation Details}
\label{sec:appendix/implementation_details}

Table~\ref{tab:hyperparameters} shows the details of hyperparameter settings in our experiments.
Our implementation was based on PyTorch~2.1.0, and we used pre-trained models from Hugging Face Transformers~4.30.0.
All model training and evaluation in our experiments were conducted with the half-precision (BFloat16) on a single RTX 3090 (24GB), RTX A6000 (48GB), or A100 PCIe (80GB) according to the required GPU memory size for each experiment.

\begin{table}[ht!]
    \centering
    \small
    \begin{adjustbox}{width=\linewidth}
    \begin{tabular}{lc}
        \toprule
        \multicolumn{2}{c}{\textbf{Pre-training settings of DiLM}} \\
        \midrule
        Optimizer & AdamW \\
        Learning rate & $1.0\times 10^{-5}$\\
        Learning rate scheduler & \begin{tabular}{c}Linear warm-up and\\cosine annealing\end{tabular}\\
        Warmup ratio & $0.05$ \\
        Waight decay & $0.01$ \\
        Gradient clipping & $1.0$ \\
        Dropout ratio & $0.1$ \\
        \# of training steps & $80{,}000$ \\
        Batch size & $64$ \\
        \midrule
        \multicolumn{2}{c}{\textbf{Fine-tuning settings of DiLM}} \\
        \midrule
        Optimizer & AdamW \\
        Learning rate & $3.0\times10^{-7}$ \\
        Learning rate scheduler & \begin{tabular}{c}Linear warm-up and\\cosine annealing\end{tabular}\\
        Warmup ratio & $0.05$ \\
        Waight decay & $0.01$ \\
        Gradient clipping & $1.0$ \\
        Dropout ratio & $0.1$ \\
        \# of outer loop ($S$) & $20{,}000$ \\
        \# of inner loop ($T$) & $10$ \\
        \# of learner updating steps ($K$) & $20$ \\
        Batch size of real samples ($M$) & $200$ \\
        Batch size of synthetic samples ($N$) & $64$ \\
        Generation interval ($I_\text{int}$) & $200$ \\
        \midrule
        \multicolumn{2}{c}{\textbf{Learner training settings for evaluation}} \\
        \midrule
        Oprimizer & AdamW \\
        Learning rate & $1.0\times10^{-4}$ \\
        Learning rate scheduler & \begin{tabular}{c}Linear warm-up and\\cosine annealing\end{tabular}\\
        Warmup ratio & $0.5$ \\
        Waight decay & $0.01$ \\
        Gradient clipping & $1.0$ \\
        Dropout ratio & $0.1$ \\
        \# of training steps & $200$ \\
        Batch size & $64$ \\
        \bottomrule
    \end{tabular}
    \end{adjustbox}
    \caption{Hyperparameter settings in our experiments}
    \label{tab:hyperparameters}
\end{table}

\section{Results for Cross-model Generalization}
\label{sec:appendix/additional_results}

Tables~\ref{tab:cross_model_generalization_dpc_5} and \ref{tab:cross_model_generalization_dpc_10} show the cross-model generalization performances with DPC=5,10 settings.
As in the setting of DPC=20 in Table~\ref{tab:cross_model_generalization}, DiLM also performed well in training different models than the source model.

\newcolumntype{Y}{>{\centering\arraybackslash}p{4.5em}}
\begin{table}[t]
    % \tabcolsep = 5pt
    % \small
    \centering
    \begin{adjustbox}{width=\linewidth}
    \begin{tabular}{cl*3Y}
        \toprule
        \textbf{Dataset} & \textbf{Model}  & \textbf{Random} & \textbf{K-centers} & \textbf{DiLM} \\
        \midrule
        \multirow{4}{*}{SST-2} 
        & BERT$_\text{BASE}$ (\textbf{S}) & $58.1{\pm 5.2}$ & $70.8{\pm 4.1}$ & $72.5{\pm 5.9}^*$ \\
        \cmidrule{2-5}
        & RoBERTa$_\text{BASE}$ & $60.6{\pm 7.6}$ & $74.2{\pm 4.9}$ & \cellcolor{yellow-green!25}$\mathbf{75.1{\pm 4.6}}$ \\
        & BERT$_\text{LARGE}$ & $60.4{\pm 8.4}$ & $70.0{\pm 8.2}$ & \cellcolor{yellow-green!25}$\mathbf{73.7{\pm 8.4}^*}$ \\
        & XLNet$_\text{BASE}$ & $57.0{\pm 5.5}$ & $66.4{\pm 5.0}$ & \cellcolor{yellow-green!25}$\mathbf{69.5{\pm 6.6}^*}$ \\
        \midrule
        \multirow{4}{*}{QQP}
        & BERT$_\text{BASE}$ (\textbf{S}) & $51.5{\pm 5.6}$ & $60.7{\pm 3.8}$ & $58.8{\pm 5.2}$ \\
        \cmidrule{2-5}
        & RoBERTa$_\text{BASE}$ & $52.5{\pm 6.0}$ & $\mathbf{63.9{\pm 3.3}}$ & $62.4{\pm 3.7}$ \\
        & BERT$_\text{LARGE}$ & $53.3{\pm 6.7}$ & $58.3{\pm 5.8}$ & \cellcolor{yellow-green!25} $\mathbf{58.8{\pm 5.7}}$ \\
        & XLNet$_\text{BASE}$ & $52.6{\pm 5.2}$ & $\mathbf{62.6{\pm 3.1}}$ & $60.2{\pm 4.6}$ \\
        \midrule
        \multirow{4}{*}{MNLI-m}  
        % & BERT$_\text{BASE}$ (\textbf{S}) & $35.6{\pm 2.1}$ & $36.2{\pm 2.4}$ & \cellcolor{yellow-green!25} $\mathbf{39.7{\pm 2.7}}$ \\
        & BERT$_\text{BASE}$ (\textbf{S}) & $35.6{\pm 2.1}$ & $36.2{\pm 2.4}$ & $39.7{\pm 2.7}^*$ \\
        \cmidrule{2-5}
        & RoBERTa$_\text{BASE}$ & $35.8{\pm 2.1}$ & $37.4{\pm 2.1}$ & \cellcolor{yellow-green!25}$\mathbf{38.8{\pm 3.0}^*}$ \\
        & BERT$_\text{LARGE}$ & $36.9{\pm 2.8}$ & $37.4{\pm 2.9}$ & \cellcolor{yellow-green!25}$\mathbf{41.5{\pm 3.7}^*}$ \\
        & XLNet$_\text{BASE}$ & $35.4{\pm 1.4}$ & $37.0{\pm 1.5}$ & \cellcolor{yellow-green!25}$\mathbf{37.3{\pm 1.9}}$ \\
        \bottomrule
    \end{tabular}
    \end{adjustbox}
    \caption{Cross-model generalization performance for the setting of DPC=5. (\textbf{S}) indicates the source model for gradient matching of DiLM and feature extractor for K-centers.}
    \label{tab:cross_model_generalization_dpc_5}
\end{table}

\begin{table}[t]
    % \small
    \centering
    \begin{adjustbox}{width=\linewidth}
    \begin{tabular}{cl*3Y}
    \toprule
    \textbf{Dataset} & \textbf{Model}  & \textbf{Random} & \textbf{K-centers} & \textbf{DiLM} \\
    \midrule
    % DPC=10
    \multirow{4}{*}{SST2}
    & BERT$_\text{BASE}$ (\textbf{S}) & $64.3{\pm 7.4}$ & $75.9{\pm 4.7}$ & $76.3{\pm 4.6}$ \\
    \cmidrule{2-5}
    & RoBERTa$_\text{BASE}$ & $68.6{\pm 7.1}$ & $74.6{\pm 5.6}$ & \cellcolor{yellow-green!25}$\mathbf{77.1{\pm 4.1}^*}$ \\
    & BERT$_\text{LARGE}$ & $67.2{\pm 8.5}$ & $76.6{\pm 8.4}$ & \cellcolor{yellow-green!25}$\mathbf{79.2{\pm 7.8}^*}$ \\
    & XLNet$_\text{BASE}$ & $63.7{\pm 7.5}$ & $68.0{\pm 6.1}$ & \cellcolor{yellow-green!25}$\mathbf{74.2{\pm 4.9}^*}$ \\
    \midrule
    \multirow{4}{*}{QQP}
    & BERT$_\text{BASE}$ (\textbf{S}) & $56.0{\pm 4.8}$ & $60.9{\pm 3.1}$ & $62.2{\pm 3.3}^*$ \\
    \cmidrule{2-5}
    & RoBERTa$_\text{BASE}$ & $56.4{\pm 5.3}$ & $\mathbf{64.0{\pm 2.7}}$ & $63.9{\pm 4.3}$ \\
    & BERT$_\text{LARGE}$ & $53.7{\pm 8.5}$ & $59.4{\pm 5.6}$ & \cellcolor{yellow-green!25}$\mathbf{60.6{\pm 7.5}}$ \\
    & XLNet$_\text{BASE}$ & $55.0{\pm 4.5}$ & $61.4{\pm 3.2}$ & \cellcolor{yellow-green!25}$\mathbf{62.8{\pm 2.2}^*}$ \\
    \midrule
    \multirow{4}{*}{MNLI-m}
    & BERT$_\text{BASE}$ (\textbf{S}) & $37.7{\pm 2.6}$ & $41.8{\pm 3.2}$ & $44.8{\pm 3.1}^*$ \\
    \cmidrule{2-5}
    & RoBERTa$_\text{BASE}$ & $37.1{\pm 2.2}$ & $\mathbf{42.1{\pm 2.6}}$ & $40.9{\pm 2.6}^*$ \\
    & BERT$_\text{LARGE}$ & $39.7{\pm 3.6}$ & $43.4{\pm 4.4}$ & \cellcolor{yellow-green!25}$\mathbf{45.4{\pm 4.1}^*}$ \\
    & XLNet$_\text{BASE}$ & $37.0{\pm 1.4}$ & $\mathbf{41.5{\pm 2.6}^*}$ & $40.6{\pm 1.9}$ \\
    \bottomrule
    \end{tabular}
    \end{adjustbox}
    \caption{Cross-model generalization performance for the setting of DPC=10. (\textbf{S}) indicates the source model for gradient matching of DiLM and feature extractor for K-centers.}
    \label{tab:cross_model_generalization_dpc_10}
\end{table}

% \subsection{Hyperparameter sensitivity}
% for the \# of loop (S, T, K) in DiLM

\section{Distilled Synthetic Data Examples}
\label{sec:appendix/examples}

We gave examples of distilled synthetic samples from DiLM in Tables~\ref{tab:example_sst2}, \ref{tab:example_qqp}, and \ref{tab:example_mnli}. 
Generated synthetic examples with DiLM were interpretable and seem to represent the tasks of the original training dataset. 

\begin{table*}
    \small
    \centering
    \begin{adjustbox}{width=\linewidth}
    \begin{tabular}{lp{44em}}
    \toprule
    \textbf{Label} & \textbf{Sentence} \\
    \midrule
    \multirow{5}{*}[-12pt]{negative} & is too amateurishly square to work as storytelling, and the ensemble cast lacks depth and resonance. \\
    \cmidrule{2-2}
    & is so lousy that you can not enjoy it \\
    \cmidrule{2-2}
    & incredibly lifeless, with the lack-of-attention span \\
    \cmidrule{2-2}
    & the script's contrived, lame screenplay and listless direction are just the ticket cost. \\
    \cmidrule{2-2}
    & a cheap scam that only weak claims to dramatic impact and creepy-crawly humor. \\
    \midrule
    \multirow{5}{*}[-20pt]{positive} & is a wonderous accomplishment of veracity and narrative grace. \\
    \cmidrule{2-2}
    & very best \\
    \cmidrule{2-2}
    & a fully realized story with keen insights into parapsychological phenomena and the soulful nuances of the grieving process \\
    \cmidrule{2-2}
    & it one of the best-sustained ideas i have ever seen on the screen. \\
    \cmidrule{2-2}
    & a surprisingly sweet, tender drama that does a superb job contrasting the sleekness of the film's present with the playful paranoia of the film's past. \\
    \bottomrule
    \end{tabular}
    \end{adjustbox}
    \caption{Distilled synthetic samples for SST-2 with DPC=5}
    \label{tab:example_sst2}
\end{table*}

\begin{table*}
    \small
    \centering
    \begin{adjustbox}{width=\linewidth}
    \begin{tabular}{lp{22em}p{22em}}
    \toprule
    \textbf{Label} & \textbf{Question~1} & \textbf{Question~2} \\
    \midrule
    \multirow{5}{*}[-35pt]{not duplicate} 
    & Why should I write a good backmatter for an international conference? & Where can I study internationally on business logic? \\
    \cmidrule{2-3}
    & How long does it take you to learn the German language? & How long does it take to learn the English language? \\
    \cmidrule{2-3}
    & What are some unexpected things first-time visitors to Colombia notice? & What are some unexpected things first-time visitors to Canada notice? \\
    \cmidrule{2-3}
    & Why is red in PFUS something I can't see when I tap PFUS? & Did one have a chance to see one of the real masterpieces being played by Richard Bachardo in MS Dhoni Cricket: Live Streaming, in the Permanent XI Test Center at Mumbai? \\
    \cmidrule{2-3}
    & How does digital gatekeeper disable ads on a WiFi band? & How can I enable\/disable my WiFi network on my HTC phone? \\
    \midrule
    \multirow{5}{*}[-18pt]{duplicate} 
    & How do I recover my Gmail account after recovery? & How do I recover my Gmail account from recovery? \\
    \cmidrule{2-3}
    & How do you prevent hair loss without touching hair? & How do I prevent hair loss without touching hair? \\
    \cmidrule{2-3}
    & How do I get successful in C.E.? & How can I get successful in C.E.? \\
    \cmidrule{2-3}
    & What is the best word or link you use to explain the meaning of a certain book to a friend? & What is the best word or link you use to explain the meaning of a certain book to a friend? \\
    \cmidrule{2-3}
    & How will the ban of Rs 500 and Rs 1000 notes affect Indian economy? & How will the 500 and 1000 rupee notes ban affect the Indian economy? \\
    \bottomrule
    \end{tabular}
    \end{adjustbox}
    \caption{Distilled synthetic samples for QQP with DPC=5}
    \label{tab:example_qqp}
\end{table*}

\begin{table*}
    \small
    \centering
    \begin{adjustbox}{width=\linewidth}
    \begin{tabular}{lp{22em}p{22em}}
    \toprule
    \textbf{Label} & \textbf{Premise} & \textbf{Hypothesis} \\
    \midrule
    \multirow{5}{*}[-90pt]{entailment}
    & Guess we are all here, friends. & We were all here, friends. \\
    \cmidrule{2-3}
    & The costs to the Service, often estimated to be between \$100 and \$150 million, will be higher because of the reduced volume of post-1991 pleadings by six states and 28 other states requiring service members to produce basic records electronically. & Costs to the Service are higher because of reduced volume of post-1991 pleadings by six states and 28 other states requiring service members to produce basic records electronically. \\
    \cmidrule{2-3}
    & uh-huh is that right because like i say a lot of people tell me we could make it cheaper if we wanted but we didn't i mean our family life is just so far so far that & It seems that a lot of people tell me that it could be cheaper if we wanted but we don't really think we could make it cheaper. \\
    \cmidrule{2-3}
    & However, the CEF report suggested that some of the following could serve to reduce the burden on small entities with federally or nonfederal support for compliance with the rule and to minimize the number of affected entities receiving small reductions of federal payments. & Some things could be considered part of the CEF report for reducing burdens on small entities. \\
    \cmidrule{2-3}
    & If you are a casino business owner looking to expand your profits, opportunities and experiences, or even to retain some intellectual property you acquired during your travels in other countries, it is best to visit Cancio, Parnell's (National Cancia) resort in Montego Bay, where prices and travel policies range from a very reasonable \$50. & The casino has plenty of opportunities you can expand your profits with in Cancio, Parnell's resort. \\
    \midrule
    \multirow{5}{*}[-40pt]{neutral}
    & oh in that case you have to give them uh six months to come and you know and let them go on & They don't have to get their first six months if they return. \\
    \cmidrule{2-3}
    & This is highly valued nationally because of its steeply pro-retirement payment culture, which is perceived as a great success rate by the profession and outside of its area of employment, particularly among the field's young professionals. & Out of all the fields in the population, it is highly valued by the professional community because it provides confidence that the community will care more about its growth. \\
    \cmidrule{2-3}
    & yeah right now i i still wish they were a little more & The idea of having people tell us what to do is good for their business and prospects. \\
    \cmidrule{2-3}
    & In fact, there is one wonder why Republican leaders are afraid to mention his name. & Republican leaders are not afraid of his name because he is in need of attention. \\
    \cmidrule{2-3}
    & To me, it's an excellent system. & I think it could be a good system for a number of reasons. \\
    \midrule
    \multirow{5}{*}[-95pt]{contradiction}
    & yeah well you know i can't i can't i know sometimes i just i'll remember remembering for once the former minister might be sympathetic to some of the Serbian government cases that they might say well there's no way out um no matter what their approach to the possibility of a peace dividend a lot of people i think i think are are willing to compromise and and to stand up and say who's right and who's wrong and i think it's a good idea and & I can't recall the minister's views on different Serbian government cases. \\
    \cmidrule{2-3}
    & I suppose you could say, if it were not for the gleam of light in the hour of your death-boom, that the fatal effects were of a furtive rather than a ferocious nature? & I don't think you could confirm it is a furtive either. \\
    \cmidrule{2-3}
    & i think something has to change there & They have no plans at all to change. \\
    \cmidrule{2-3}
    & The revisions take into account the range of factors that varying units of measure represent when evaluating new disclosure requirements and when determining whether it should be possible to offer various types of similar products for different reasons. & The revisions go against the current practice and do not consider whether it should be possible to offer different types of similar products for different reasons. \\
    \cmidrule{2-3}
    & yeah i uh i uh i don't think there's that's a bad place to live in some part of the world and do everything else that it's really not because people have gotten up in arms but it's all it's all a lot of money to run a very very wealthy individual home & I don't think we should be buying a very wealthy home in an undeveloped area in the developed world. \\
    \bottomrule
    \end{tabular}
    \end{adjustbox}
    \caption{Distilled synthetic samples for MNLI-m with DPC=5}
    \label{tab:example_mnli}
\end{table*}

\end{document}